\documentclass[runningheads,envcountsame,a4paper]{llncs}
\usepackage[pdftex]{graphicx}
\usepackage{epstopdf}
\usepackage{amssymb}
\usepackage{amsmath}
\usepackage{amsfonts}
\usepackage[hyphens]{url}
\usepackage{multirow}
\usepackage{algorithm}
\usepackage{booktabs}
\usepackage{algpseudocode}
\usepackage{color}
\usepackage{float}
\usepackage{subfig}
\usepackage[misc]{ifsym}
\usepackage{bm}
\usepackage{thmtools,thm-restate}
\newcommand{\norm}[1]{\left\lVert#1\right\rVert}
\begin{document}
%
\title{On Generalization of Graph Autoencoders with Adversarial Training }
%
%
\author{Tianjin Huang \Letter \inst{1}  \and
Yulong Pei\inst{1} \and
Vlado Menkovski\inst{1} \and
Mykola Pechenizkiy \inst{1}}

\authorrunning{T. Huang et al.}

\institute{Department of Mathematics and Computer Science, Eindhoven University of Technology, 5600 MB Eindhoven, the Netherlands\\
\email{\{t.huang,y.pei.1,v.menkovski, m.pechenizkiy\}@tue.nl}}

\maketitle              
\begin{abstract}
Adversarial training is an approach for increasing model's resilience against adversarial perturbations. Such approaches have been demonstrated to result in models with feature representations that generalize better. However, limited works have been done on adversarial training of models on graph data. In this paper, we raise such a question -- does adversarial training improve the generalization of graph representations. We formulate $L_{2}$ and $L_{\infty}$ versions of adversarial training in two powerful node embedding methods: graph autoencoder (GAE) and variational graph autoencoder (VGAE).  We conduct extensive experiments on three main applications, i.e.\ link prediction, node clustering, graph anomaly detection of GAE and VGAE, and demonstrate that both $L_{2}$ and $L_{\infty}$ adversarial training boost the generalization of GAE and VGAE.
\keywords{Graph Autoencoders  \and Variational Graph Autoencoders \and Adversarial Training \and Node Embedding \and Generalization.}
\end{abstract}
\section{Introduction}

Networks are ubiquitous in a plenty of real-world applications and they contain relationships between entities and  attributes of entities. 
Modeling such data is challenging due to its non-Euclidean characteristic. Recently, graph embedding that converts graph data into low dimensional feature space has emerged as a popular method to model graph data, For example, DeepWalk~\cite{perozzi2014deepwalk}, node2vec~\cite{grover2016node2vec} and LINE~\cite{tang2015line} learn graph embedding by extracting patterns from the graph. Graph Convolutions Networks (GCNs)~\cite{kipf2016semi} learn graph embedding by repeated multiplication of normalized adjacency matrix and feature matrix. In particular, graph autoencoder (GAE)~\cite{kipf2016variational,tian2014learning,wang2016structural} and variational graph autoencoder (VGAE)~\cite{kipf2016variational} have been shown to be powerful node embedding methods as unsupervised learning.  They have been applied to many machine learning tasks, e.g. node clustering~\cite{salha2020simple,tian2014learning,shi2020effective}, link prediction~\cite{schlichtkrull2018modeling,kipf2016variational}, graph anomaly detection~\cite{pei2020resgcn,ding2019deep} and etc.  

Adversarial training is an approach for increasing model's resilience against adversarial perturbations by including adversarial examples in the training set~\cite{madry2017towards}. 
Several recent studies demonstrate that adversarial training improves feature representations leading to better performance for downstream tasks~\cite{utrera2020adversarially,salman2020adversarially}. However, little work in this direction has been done for GAE and VGAE. Besides, real-world graphs are usually highly noisy and incomplete, which may lead to a sub-optimal results for standard trained models~\cite{yu2019graph}. Therefore, we are interested to seek answers to the following two questions: 
\begin{itemize}
    \item \emph{Does adversarial training improve generalization, i.e.\ the performance in applications of node embeddings learned by GAE and VGAE?} 
    \item \emph{Which factors influence this improvement?}
\end{itemize}
In order to answer the first question above, we firstly formulate $L_{2}$ and $L_{\infty}$ adversarial training for GAE and VGAE. Then, we select three main tasks of VGAE and GAE: link prediction, node clustering and graph anomaly detection for evaluating the generalization performance brought by adversarial training.
Besides, we empirically explore which factors affect the generalization performance brought by adversarial training.

\textbf{Contributions:} To the best of our knowledge, we are the first to explore  generalization for GAE and VGAE using adversarial training. We formulate $L_{2}$ and $L_{\infty}$ adversarial training, and empirically demonstrate that  both $L_{2}$ and $L_{\infty}$ adversarial training boost the generalization with a large margin for the node embeddings learned by GAE and VGAE. An additional interesting finding is that the generalization performance of the proposed adversarial training is more sensitive to attributes perturbation than adjacency matrix perturbation and not sensitive to the degree of nodes. 

\section{Related Work}
Adversarial training has been extensively studied in images. It has been important issues to explore whether adversarial training can help generalization. Tsipras et al.~\cite{tsipras2018robustness} illustrates that adversarial robustness could conflict with model's generalization by a designed simple task. However, Stutz et al.~\cite{stutz2019disentangling} demonstrates that adversarial training with on-manifold adversarial examples helps the generalization. Besides, Salman et al.~\cite{salman2020adversarially} and Utrera et al.~\cite{utrera2020adversarially} show that the latent features learned by adversarial training are improved and boost the performance of their downstream tasks.  

Recently, few works bring adversarial training in graph  
data. Deng, Dong and Zhu~\cite{deng2019batch} and Sun et al.~\cite{sun2019virtual} propose virtual graph adversarial training to promote the smoothness of model. Feng et al.~\cite{feng2019graph} propose graph adversarial training by inducing dynamical regularization. Dai et al.~\cite{dai2019adversarial} formulate an interpretable adversarial training for DeepWalk.  Jin and Zhang~\cite{jin2019latent} introduce latent adversarial training for GCN, which train GCN based on the adversarial perturbed output of the first layer. Besides, several studies explored adversarial training based on adversarial perturbed edges for graph data~\cite{xu2019topology,chen2019can,wang2019graphdefense}. Among these works,part of studies pay attention to achieving model's robustness while ignoring the effect of generalization~\cite{xu2019topology,jin2019latent,chen2019can,wang2019graphdefense,dai2019adversarial} and the others simply utilize perturbations on nodal attributes while not explore the effect of perturbation on edges~\cite{deng2019batch,sun2019virtual,feng2019graph}. The difference between these works and ours is two-fold: 
(1) We extend both $L_{\infty}$ and $L_{2}$ adversarial training for graph models while the previous studies only explore $L_{2}$ adversarial training. (2) We focus on the generalization effect brought by adversarial training for unsupervised deep learning graph models, i.e.\ GAE and VGAE while most of the previous studies focus on adversarial robustness for supervised/semi-supervised models.   

\section{Preliminaries}
We first summarize some notations and definitions used in this paper. Following the commonly used notations, we use bold uppercase characters for matrices, e.g.\ $\bm{X}$, bold lowercase characters for vectors, e.g.\ $\bm{b}$, and normal lowercase characters for scalars, e.g. $c$. The $i^{th}$ row of a matrix $\bm{A}$ is denoted by $\bm{A}_{i,:}$ and $(i,j)^{th}$ element of matrix $\bm{A}$ is denoted as $\bm{A}_{i,j}$. The $i^{th}$ row of a matrix $\bm{X}$ is denoted by $\bm{x}_{i}$. We use $\mathbf{KL}$ for Kullback-Leibler divergence. 

We consider an attributed network $\mathcal{G}=\{V,E,\bm{X}\}$ with $|V|=n$ nodes, $|E|=m$ edges and $\bm{X}$ node attributed matrix. $\bm{A}$ is the binary adjacency matrix of $\mathcal{G}$.
\subsection{Graph Autoencoders}
Graph autoencoders is a kind of unsupervised learning models on graph-structure data~\cite{kipf2016variational}, which aim at learning low dimensional representations for each node by reconstructing inputs. It has been demonstrated to achieve competitive results in multiple tasks, e.g. link prediction~\cite{salha2020simple,kipf2016variational,salha2019gravity}, node clustering~\cite{salha2020simple,tian2014learning,shi2020effective}, graph anomaly detection~\cite{ding2019deep,pei2020resgcn}. 
Generally, graph autoencoder consists of a graph convolutional network for encoder and an inner product for decoder~\cite{kipf2016variational}. Formally, it can be expressed as follows:
\begin{align}
    \bm{Z}=GCN(\bm{A},\bm{X}) \\
    \hat{\bm{A}}=\sigma(\bm{Z}\bm{Z}^{T}),
\end{align}
where $\sigma$ is the sigmoid function, $GCN$ is a graph convolutional network, $\bm{Z}$ is the learned low dimensional representations and $\hat{\bm{A}}$ is the reconstructed adjacency matrix.

During the training phase, the parameters will be updated by minimizing the reconstruction loss. Usually, the reconstruction loss is expressed as cross-entropy loss between $\bm{A}$ and $\hat{\bm{A}}$~\cite{kipf2016variational}:
\begin{align}
    \mathcal{L}^{ae}=-\frac{1}{n^2}\sum_{(i,j)\in V\times V} \Bigg[ \bm{A}_{i,j} log\hat{\bm{A}}_{i,j}+(1-\bm{A}_{i,j})log(1-\hat{\bm{A}}_{i,j} )\Bigg].
    \label{gae_loss}
\end{align}

\subsection{Variational Graph Autoencoders}
Kipf and Welling~\cite{kipf2016variational} introduced variational graph autoencoder (VGAE) which is a probabilistic model. VGAE is consisted of inference model and generative model. In their approach, the inference model, i.e.\ corresponding to the encoder of VGAE, is expressed as follows:
\begin{align}
    q(\bm{Z}|\bm{X},\bm{A})=\prod_{i=1}^{n}q(\bm{z}_{i}|\bm{X},\bm{A}),\; with\; q(\bm{z}_i|\bm{X},\bm{A})=\mathcal{N}(\bm{z}_i|\bm{\mu}_i,diag(\bm{\sigma}_{i}^{2})),
\end{align}
where $\bm{\mu_i}$ and $\bm{\sigma_{i}}$ are learned by a graph neural network respectively. That is, $\bm{\mu}=GCN_{\mu}(\bm{X},\bm{A})$ and $log \bm{\sigma}=GCN_{\delta}(\bm{X},\bm{A})$, with $\bm{\mu}$ is the matrix of stacking  vectors $\bm{\mu_{i}}$; likewise, $\bm{\sigma}$ is the matrix of stacking vectors $\delta_{i}$. 

The generative model, i.e.\ corresponding to the decoder of autoencoder, is designed as an inner product between latent variables $\bm{Z}$, which is formally expressed as follows:
\begin{align}
    p(\bm{A}|\bm{Z})=\prod_{i=1}^{n}\prod_{j=1}^{n}p(\bm{A}_{i,j}|\bm{z}_i,\bm{z}_j),\; with\; p(\bm{A}_{i,j}=1|\bm{z}_i,\bm{z}_j)=\sigma(\bm{z}_{i}^T\bm{z}_j).
\end{align}

During the training phase, the parameters will be updated by minimizing the the variational lower bound $\mathcal{L}^{vae}$:
\begin{align}
    \mathcal{L}^{vae}=\mathbf{E}_{q(\bm{Z}|\bm{X},\bm{A})} [log p(\bm{A}|\bm{Z})]-\mathbf{KL}(q(\bm{Z}|\bm{X},\bm{A})||p(\bm{Z})),
    \label{vgae_loss}
\end{align}
where a Gaussian prior is adopted for $p(\bm{Z})=\prod_{i}p(\bm{z}_i)=\prod_{i} \mathcal{N}(\mathbf{z}_i|0,\mathbf{I})$.

\subsection{Adversarial Training}
By now, multiple variants of adversarial training has been proposed and most of them are built on supervised learning and Euclidean data, e.g.\ FGSM-adversarial training~\cite{goodfellow2014explaining}, PGD-adversarial training~\cite{madry2017towards}, Trades~\cite{zhang2019theoretically}, MART~\cite{wang2019improving} and etc. Here we introduce Trades that will be extended to GAE and VGAE settings in Section~\ref{GAT}. Trades~\cite{zhang2019theoretically} separates loss function into two terms:1) Cross-Entropy Loss for achieving natural accuracy; 2) Kullback-Leibler divergence for achieving adversarial robustness. Formally, given inputs $(X,Y)$, it can be expressed as follows~\cite{zhang2019theoretically}:
\begin{align}
    \min_{\theta} \mathbf{E}_{(X,Y)} [L(f_{\theta}(X),Y)+\lambda \cdot \mathbf{KL}(P(Y|X')||P(Y|X))],
    \label{trades}
\end{align}
where $f_{\theta}$ is a supervised model, $X'$ is the adversarial examples that maximize $\mathbf{KL}$ divergence and $P(Y|X)$ is the output probability after softmax. $\lambda$ is a tunable hyperparameter and it controls the strength of the $\mathbf{KL}$ regularization term.

\section{Graph Adversarial Training}\label{GAT}
In this section, we formulate $L_{2}$ and $L_{\infty}$ adversarial training for GAE and VGAE respectively. 

\subsection{Adversarial Training in Graph Autoencoder}
Considering that: (1) the inputs of GAE contains adjacency matrix and attributes, (2) the latent representation $\bm{Z}$ is expected to be invariant to the input perturbation, we reformulate the loss function in Eq.~\ref{gae_loss} as follows:
\begin{align}
    &\min_{\theta}  \mathcal{L}^{ae}+\lambda \cdot \mathbf{KL}(P(\bm{Z}|\bm{A'},\bm{X'})||P(\bm{Z}|\bm{A},\bm{X})) \\
     & \bm{X}'=arg \max_{\norm{\bm{X}'-\bm{X}} \leq \epsilon} \mathcal{L}^{ae}(\bm{A},\bm{X}),\; \bm{A}'=arg \max_{\norm{\bm{A}'-\bm{A}} \leq \epsilon} \mathcal{L}^{ae}(\bm{A},\bm{X})
\label{AE_obj}
\end{align}
where $\bm{A}'$ is the adversarial perturbed adjacency matrix and $\bm{X}'$ is the adversarial perturbed attributes. Here the important question is how to  generate the perturbed adjacency matrix $\bm{A}'$ and attributes $\bm{X}'$ in Eq.~\ref{AE_obj}.

\textbf{Attributes Perturbation $\bm{X}'$}.
We generate the perturbed $X'$ by projection gradient descent (PGD)~\cite{madry2017towards}. We denote total steps as $T$.  

For  $\bm{X}'$ bounded by $L_{2}$ norm ball, the perturbed data in $t$-th step $\bm{X}^{t}$ is expressed as follows:
\begin{align}
    &\bm{X}^t=\prod_{\mathcal{B}(\bm{X},\epsilon\norm{X}_{2})} (\bm{X}^{t-1}+\alpha \cdot g \cdot \norm{X}_{2}/\norm{g}_{2})\\
     &g=\nabla_{\bm{X}^{t-1}} \mathcal{L}^{ae}(\bm{A},\bm{X}^{t-1}) 
     \label{att_perturb_l2}
\end{align}
where $\prod$ is the projection operator and $\mathcal{B}(\bm{X},\epsilon\norm{X}_{2})$ is the $L_{2}$ norm ball of nodal attributes $\bm{x}_{i}:\{\bm{x}'_{i}:\norm{\bm{x}'_{i}-\bm{x}_{i}}_{2} \leq \epsilon\norm{\bm{x}_{i}}_{2}\}$. 

For $\bm{X}'$ bounded by $L_{\infty}$ norm ball, the perturbed data in $t$-th step $\bm{X}^{t}$ is expressed as follows:
\begin{align}
      &\bm{X}^t=\prod_{\mathcal{B}(\bm{X},\epsilon)} (\bm{X}^{t-1}+\alpha \cdot g)\\
     &g=sgn(\nabla_{\bm{X}^{t-1}} \mathcal{L}^{ae}(\bm{A},\bm{X}^{t-1})), 
     \label{att_perturb_linf}
\end{align}
where $\mathcal{B}(\bm{X},\epsilon)$ is the $L_{\infty}$ norm ball of nodal attributes $\bm{x}_{i}:\{\bm{x}'_{i}:\norm{\bm{x}'_{i}-\bm{x}_{i}}_{\infty} \leq \epsilon \}$ and $sgn(\cdot)$ is the sign function.

\textbf{Adjacency Matrix Perturbation $\bm{A}'$}. Adjacency matrix perturbation includes two-fold:(1) perturb node connections, i.e.\ Adding or dropping edges, (2) perturb the strength of information flow between nodes, i.e.\ the strength of correlation between nodes. Here we choose to perturb the strength of information flow between nodes and leave the perturb of node connections for future work. Specifically, we add weight for each edge and change these weights in order to perturb the strength of information flow. Formally, given the adjacency matrix $A$, the weighted adjacency matrix $\Tilde{\bm{A}}$ is expressed as $\bm{A} \odot \bm{M}$ where the elements of $\bm{M}$ are continuous and its values are initialized as same value as $\bm{A}$. $\odot$ denotes the element-wise product. Formally, $\bm{A}'$ is expressed as follows:
\begin{align}
    &\bm{M}'=arg \max_{\norm{\bm{M}'-\bm{M}} \leq \epsilon} \mathcal{L}^{ae}(\Tilde{\bm{A}},\bm{X}) \\
    &\bm{A}'=\bm{A}\odot \bm{M}'.
\end{align}
For $\bm{A}'$ bounded by $L_{2}$ norm ball, the perturbed data in $t$-th step $\bm{A}^{t}$ is expressed as follows:
\begin{align}
    &g=\nabla_{\bm{M}^{t-1}} \mathcal{L}^{ae}(\Tilde{\bm{A}}^{t-1},\bm{X}) \\
    &\bm{M}^{t}=\prod_{\mathcal{B}(\bm{M},\epsilon \norm{M}_{2})} (\bm{M}^{t-1}+\alpha \cdot g \cdot \norm{M}_{2}/\norm{g}_{2}) \\
    &\bm{A}^t=\Tilde{\bm{A}}^{t}=\bm{A}\odot\bm{M}^t.
    \label{edge_perturb_l2}
\end{align}
 
For $\bm{A}'$ bounded by $L_{\infty}$ norm ball, the perturbed data in $t$-th step $\bm{A}^{t}$ is expressed as follows:
\begin{align}
    &g=sgn(\nabla_{\bm{M}^{t-1}} \mathcal{L}^{ae}(\Tilde{\bm{A}}^{t-1},\bm{X}))  \\
    &\bm{M}^{t}=\prod_{\mathcal{B}(\bm{M},\epsilon)} (\bm{M}^{t-1}+\alpha \cdot g) \\
    &\bm{A}^t=\Tilde{\bm{A}}^{t}=\bm{A}\odot\bm{M}^t.
    \label{edge_perturb_linf}
\end{align}

\subsection{Adversarial Training in Variational Graph  Autoencoder}
Similarly to GAE, we reformulate the loss function for training VGAE (Eq.~\ref{vgae_loss}) as follows:
\begin{align}
    &\min_{\theta}  \mathcal{L}^{vae}+\lambda \cdot \mathbf{KL}(P(\bm{Z}|\bm{A'},\bm{X'})||P(\bm{Z}|\bm{A},\bm{X})) \\
     & \bm{X}'=arg \max_{\norm{\bm{X}'-\bm{X}} \leq \epsilon} \mathcal{L}^{vae}(\bm{A},\bm{X}),\; \bm{A}'=arg \max_{\norm{\bm{A}'-\bm{A}} \leq \epsilon} \mathcal{L}^{vae}(\bm{A},\bm{X})
\label{vAE_obj}
\end{align}
We generate $\bm{A}'$ and $\bm{X}'$ exactly the same way as with GAE (replacing $\mathcal{L}^{ae}$ with $\mathcal{L}^{vae}$ in Eq.~10-21.) 

For convenience, we abbreviate $L_{2}$ and $L_{\infty}$ adversarial training  as AT-2 and AT-Linf respectively in the following tables and figures where $L_{2}$/$L_{\infty}$ denote both attributes and adjacency matrix perturbation are bounded by $L_{2}$/$L_{\infty}$ norm ball.

In practice, we train models by alternatively adding adjacency matrix perturbation and attributes perturbation~\footnote{We find that optimizing models by alternatively adding these two perturbation is better than adding these two perturbation together (See Appendix).}. 

\section{Experiments}
In this section, we present the results of  the performance evaluation of $L_{2}$ and $L_{\infty}$ adversarial training under three main applications of GAE and VGAE: link prediction, node clustering, and graph anomaly detection. Then we conduct parameter analysis experiments to explore which factors influence the performance. 

\textbf{Datasets}. We used six real-world datasets: Cora, Citeseer and PubMed for link prediction and node clustering tasks, and  BlogCatalog, ACM and Flickr for the graph anomaly detection task. The detailed descriptions of the six datasets are showed in Table~\ref{datasets}.

\textbf{Model Architecture}. All our experiments are based on the GAE/VGAE model where the encoder/inference model is consisted with a two-layer GCN by default. 
\begin{table}[!htb]
\centering
\caption{Datasets Descriptions.}\label{datasets}
\begin{tabular}{l|ccc|ccc}
\toprule[1pt]
DataSets &  Cora &Citeseer &PubMed &BlogCatalog &ACM &Flickr \\
\midrule[1pt]
\#Nodes &  2708 & 3327&19717 &5196&16484 &7575\\
\#Links &  5429& 4732&44338 &171743&71980&239738\\
\#Features & 1433 &3703&500&8189&8337&12074\\
\bottomrule[1pt]
\end{tabular}
\vspace{-0.1in}
\end{table}

\subsection{Link Prediction}\label{linkp}
\textbf{Metrics}. Following~\cite{kipf2016variational}, we use the area under a receiver operating characteristic curve (AUC) and average precision (AP) as the evaluation metric. We conduct 30 repeat experiments with random splitting datasets into 85\%, 5\% and 10\%  for training sets,  validation sets and test sets respectively. We report the mean and standard deviation values on test sets.\\
\textbf{Parameter Settings}. 
We train models on Cora and Citeseer datasets with 600 epochs, and PubMed with 800 epochs. All models are optimized with Adam optimizer and 0.01 learning rate. The $\lambda$ is set to $4$.  For attributes perturbation, the $\epsilon$ is set to 3e-1 and 1e-3 on Citeseer and Cora, 1 and 5e-3 on PubMed for $L_{2}$ and $L_{\infty}$ adversarial training respectively. For adjacency matrix perturbation, the $\epsilon$ is set to 1e-3 and 1e-1 on Citeseer and Cora, and 1e-3 and 3e-1 on PubMed for $L_{2}$ and $L_{\infty}$ adversarial training respectively. The steps $T$ is set to $1$. The $\alpha$ is set to $\frac{\epsilon}{T}$. 

For standard training GAE and VGAE, we run the official Pytorch geometric code~\footnote{\url{https://github.com/rusty1s/pytorch_geometric/blob/master/examples/autoencoder.py}} with 600 epochs for Citeseer and Cora datasets, 1000 epochs~\footnote{Considering PubMed is big graph data, we use more epochs in order to avoiding underfitting.} for PubMed dataset. Other parameters are set the same as in~\cite{kipf2016variational}.\\
\textbf{Experimental Results}.
The results are showed in Table~\ref{Res_link}. It can be seen that both $L_{2}$ and $L_{\infty}$ Adversarial trained GAE and VGAE models consistently boost their performance for both AUC and AP metrics on Cora, Citeseer and PubMed datasets. Specifically, the improvements on Cora and Citeseer dataset reaches at least 2\% for both GAE and VGAE (Table~\ref{Res_link}). The improvements on PubMed is relative small with around 0.3\%. 

\begin{table}[!htb]
\centering
\caption{Results for Link Prediction.}\label{Res_link}
\begin{tabular}{l|cccccc}
\toprule[1pt]
Methods &\multicolumn{2}{c}{Cora} &\multicolumn{2}{c}{Citeseer}&\multicolumn{2}{c}{PubMed}\\
\midrule[1pt]
& AUC (in\%)&AP (in\%)&AUC (in\%)&AP (in\%)&AUC (in\%)&AP (in\%)\\
\hline
GAE &$90.6 \pm 0.9$&$91.2 \pm 1.0$ &$88.0\pm 1.2$ &$89.2 \pm 1.0$ &$96.8 \pm 0.2$&$97.1\pm 0.2$\\
AT-L2-GAE & $\textbf{93.0}\pm 0.9$&$\textbf{93.5} \pm 0.6$ &$\textbf{92.5}\pm 0.7$&$\textbf{93.2}\pm 0.6$ &$\textbf{97.2} \pm 0.2$&$\textbf{97.4}\pm 0.2$\\
AT-Linf-GAE &$92.8 \pm 1.1$&$93.4\pm1.0$ &$92.3 \pm 0.9$ & $92.6\pm 1.1$&$96.9\pm 0.2$&$97.3 \pm 0.2$\\
\hline
VGAE &$89.8\pm 0.9$&$90.3\pm 1.0$&$86.6\pm 1.4$ &$87.6 \pm 1.3$ &$96.2\pm 0.4$& $96.3\pm 0.4$ \\
AT-L2-VGAE &$\textbf{92.8} \pm 0.6$&$\textbf{93.1}\pm 0.6$&$90.7\pm 1.1$ &$91.1 \pm 0.9$ &$\textbf{96.6}\pm0.2$&$\textbf{96.7} \pm 0.2$ \\
AT-Linf-VGAE&$92.2 \pm 1.2$ &$92.3 \pm 1.3$ &$\textbf{91.9}\pm 0.8$ &$\textbf{92.0} \pm 0.6$ &$96.5 \pm 0.2$&$96.6\pm 0.3$\\
\bottomrule[1pt]
\end{tabular}
\vspace{-0.1in}
\end{table}

\subsection{Node Clustering}
\textbf{Metrics}. Following~\cite{pan2018adversarially,xia2014robust}, we use accuracy (ACC), normalized mutual information (NMI), precision, F-score(F1) and average rand index (ARI) as our evaluation metrics. We conduct 10 repeat experiments. For each experiment, datasets are random split into training sets( 85\% edges), validation sets (5\% edges) and test sets (10\% edges). We report the mean and standard deviation values on test sets.\\
\textbf{Parameter Settings}.
We train GAE models on Cora and Citeseer datasets with 400 epochs, and PubMed dataset with 800 epochs. We train VGAE models on Cora and Citeseer datasets with 600 epochs and PubMed dataset with 800 epochs. All models are optimized by Adam optimizer with 0.01 learning rate. The $\lambda$ is set to 4. For attributes perturbation, the $\epsilon$ is set to 5e-1 and 1e-3 on both Cora and Citeseer dataset, and 1 and 5e-3 on PubMed dataset for $L_{2}$ and $L_{\infty}$ adversarial training respectively. For adjacency matrix perturbation, the $\epsilon$ is set to 1e-3 and 1e-1 on Cora and CiteSeer, 1e-3 and 3e-1 on PubMed for $L_{2}$ and $L_{\infty}$ adversarial training respectively. The steps $T$ is set to 1. The $\alpha$ is set to $\frac{\epsilon}{T}$.

Likewise, for standard GAE and VGAE, we run the official Pytorch geometric code with 400 epochs for Citeseer and Cora datasets, 800 epochs for PubMed dataset. \\
\textbf{Experimental Results}.
The results are showed in Table~\ref{res_clustering_cora}, Table~\ref{res_clustering_citeseer} and Table~\ref{res_clustering_pubmed}. It can be seen that both $L_{2}$ and $L_{\infty}$ adversarial trained models consistently outperform the standard trained models for all metrics. In particular, on Cora and Citeseer datasets, both $L_{2}$ and $L_{\infty}$ adversarial training improve the performance with large margin for all metrics, i.e. at least +5.4\% for GAE, +6.7\% for VGAE on Cora dataset (Table~\ref{res_clustering_cora}), and at least +5.8\% for GAE, +5.6\% for VGAE on Citeseer dataset (Table~\ref{res_clustering_citeseer}).  

\begin{table}[!htb]
\centering
\caption{Results for Node Clustering on Cora.}\label{res_clustering_cora}
\begin{tabular}{l|cccccc}
\toprule[1pt]
Methods &  Acc (in\%) & NMI (in\%)& F1 (in\%)&Precision (in\%)&ARI (in\%)\\
\midrule[1pt]
GAE &$61.6\pm3.4$ &$44.9\pm 2.3$&$60.8\pm3.4$&$62.5\pm3.5$ &$37.2\pm3.2$\\
AT-L2-GAE & $67.0\pm 3.0$&$50.8\pm1.7$&$66.6\pm1.7$&$69.4\pm1.7$ &$\textbf{44.1}\pm4.1$\\
AT-Linf-GAE& $\textbf{67.1}\pm 3.8$& $\textbf{51.4}\pm1.9$&$\textbf{67.5}\pm2.8$&$\textbf{70.7}\pm 2.2$ &$43.4\pm 4.3$\\
\hline
VGAE &$58.7\pm 2.7$ &$42.3\pm2.2$&$57.3\pm3.2$ &$58.8\pm3.5$ &$34.6\pm2.8$\\
AT-L2-VGAE &$\textbf{67.3}\pm3.8$ &$\textbf{50.5}\pm2.1$&$\textbf{66.1}\pm4.1$&$\textbf{67.5}\pm3.8$ &$\textbf{44.3}\pm3.3$ \\
AT-Linf-VGAE& $65.4\pm2.3$&$49.5\pm1.6$&$64.0\pm2.3$&$65.8\pm3.0$ &$42.9\pm2.8$\\
\bottomrule[1pt]
\end{tabular}
\vspace{-0.1in}
\end{table}

\begin{table}[!htb]
\centering
\caption{Results for Node Clustering  on Citeseer.}\label{res_clustering_citeseer}
\begin{tabular}{l|cccccc}
\toprule[1pt]
Methods &  Acc (in\%)& NMI (in\%)& F1 (in\%)&Precision (in\%)&ARI (in\%)\\
\midrule[1pt]
GAE &$51.8\pm2.6$&$28.0\pm1.9$&$50.6\pm3.1$&$55.1\pm3.1$&$22.8\pm2.3$\\
AT-L2-GAE &$\textbf{61.6}\pm2.3$&$36.3\pm1.4$ &$\textbf{58.8}\pm2.1$&$60.9\pm1.4 $ &$\textbf{34.6}\pm2.3$\\
AT-Linf-GAE&$60.2\pm2.8$ &$\textbf{38.0}\pm 2.3$&$57.0\pm 2.7$&$\textbf{61.1}\pm1.6$ &$34.1\pm 3.4$\\
\hline
VGAE &$53.6\pm3.5$&$28.4\pm3.3$&$51.1\pm3.8$&$53.2\pm4.1$&$26.1\pm3.5$\\
AT-L2-VGAE & $59.2\pm2.3$&$35.1\pm2.3$&$57.3\pm 2.3$&$60.4\pm 3.1$ &$33.0\pm2.4$ \\
AT-Linf-VGAE& $\textbf{60.4}\pm1.5$&$\textbf{36.5}\pm1.4$ &$\textbf{58.2}\pm1.4$&$\textbf{61.1}\pm1.4$ &$\textbf{34.7}\pm2.0$\\
\bottomrule[1pt]
\end{tabular}
\vspace{-0.1in}
\end{table}

\begin{table}[!htb]
\centering
\caption{Results for Node Clustering on PubMed.}\label{res_clustering_pubmed}
\begin{tabular}{l|cccccc}
\toprule[1pt]
Methods &  Acc (in\%)& NMI (in\%)& F1 (in\%)&Precision (in\%)&ARI (in\%)\\
\midrule[1pt]
GAE &$66.2\pm2.0$ & $27.9\pm3.7$&$65.0\pm2.3$&$68.8\pm2.2$ &$27.1\pm3.3$\\
AT-L2-GAE & $67.5\pm 2.9$&$30.4\pm 5$&$66.7\pm 3.3$&$70.2\pm 3.1$ &$28.9\pm 4.8$\\
AT-Linf-GAE&$\textbf{68.4}\pm 1.6$ & $\textbf{31.9}\pm3.2$&$\textbf{67.7}\pm1.9$&$\textbf{70.9}\pm1.8$ &$\textbf{30.2}\pm2.8$\\
\hline
VGAE &$67.5\pm2.0$ &$29.4\pm 3.2$&$66.5\pm2.2$ &$69.9\pm 2.2$ &$28.4\pm 3.2$\\
AT-L2-VGAE & $\textbf{69.8}\pm2.0$&$\textbf{33.2}\pm3.4$&$\textbf{69.4}\pm2.3$&$\textbf{71.7}\pm2.5$&$\textbf{32.5}\pm3.2$ \\
AT-Linf-VGAE&$68.5\pm1.2$&$30.7\pm2.5$ &$67.4\pm1.5$&$70.1\pm1.5$ &$30.4\pm2.0$\\
\bottomrule[1pt]
\end{tabular}
\vspace{-0.1in}
\end{table}

\subsection{Graph Anomaly Detection}
We exactly follow~\cite{ding2019deep} to conduct experiments for graph anomaly detection. In~\cite{ding2019deep}, the authors take reconstruction errors of attributes and links  as the anomaly scores. Specifically, the node with larger scores are more likely to be considered as anomalies.\\
\textbf{Model Architecture}. Different from link prediction and node clustering, the model architecture in graph anomaly detection not only contains structure reconstruction decoder, i.e. link reconstruction,  but also contains attribute reconstruction decoder. We adopt the same model architecture as in the official code of ~\cite{ding2019deep} where the encoder is consisted of two GCN layers, and the decoder of structure reconstruction decoder is consisted of a GCN layer and a InnerProduction layer, and the decoder of attributes reconstruction decoder is consisted of two GCN layers.\\  
\textbf{Metrics}. Following~\cite{ding2019deep,pei2020resgcn}, we use the area under the receiver operating characteristic curve (ROC-AUC) as the evaluation metric. \\
\textbf{Parameter Settings}.
We set the $\alpha$ in anomaly scores to $0.5$ where it balances the structure reconstruction errors and attributes reconstruction errors. We train the GAE model on Flickr, BlogCatalog and ACM datasets with 300 epochs. We set $\lambda$ to $5$. For adjacency matrix perturbation, we set $\epsilon$ to 3e-1, 5e-5 on both BlogCatalog and ACM datasets, 1e-3 and 1e-6 on Flickr dataset for $L_{\infty}$ and $L_{2}$ adversarial training respectively. For attributes perturbations, we set $\epsilon$ to 1e-3 on BlogCatalog for both $L_{\infty}$ and $L_{2}$ adversarial training, 1e-3 and 1e-2 on ACM for $L_{\infty}$ and $L_{2}$ adversarial training respectively, 5e-1 and 3e-1 on Flickr for $L_{\infty}$ and $L_{2}$ adversarial training respectively. We set steps $T$ to 1 and the $\alpha$ to $\frac{\epsilon}{T}$\\
\textbf{Anomaly Generation}. Following~\cite{ding2019deep}, we inject two kinds of anomaly by perturbing structure and nodal attributes respectively:
\begin{itemize}
    \item Structure anomalies. We randomly select $s$ nodes from the network and then make those nodes fully connected, and then all the $s$ nodes forming the clique are labeled as anomalies. $t$ cliques are generated repeatedly and totally there are $s \times t$ structural anomalies.
    
    \item Attribute anomalies. We first randomly select $s\times t$ nodes as the attribute perturbation candidates. For each selected node $v_{i}$, we randomly select another $k$ nodes from the network and calculate the Euclidean distance between $v_{i}$ and all the $k$ nodes. Then the node with largest distance is selected as $v_{j}$ and the attributes of node $v_j$ is changed to the attributes of $v_i$.
\end{itemize}
In this experiments, we set $s=15$ and $t=10,15,20$ for BlogCatalog, Flickr and ACM respectively which are the same to~\cite{ding2019deep,pei2020resgcn}.\\
\textbf{Experimental Results}.
From Table~\ref{res_anomaly}, it can be seen that both $L_{2}$ and $L_{\infty}$ adversarial training boost the performance in detecting anomalous nodes. Since adversarial training tend to learn feature representations that are less sensitive to perturbations in the inputs, we conjecture that the adversarial trained node embeddings are less influenced by the anomalous nodes, which helps the graph anomaly detection.  A similar claim are also made in image domain~\cite{salehi2020arae} where they demonstrate adversarial training of autoencoders are beneficial to novelty detection. 
\begin{table}[!htb]
\centering
\caption{Results  w.r.t. AUC (in\%) for Graph Anomaly Detection.}\label{res_anomaly}
\begin{tabular}{l|ccccc}
\toprule[1pt]
Methods &  Flickr & BlogCatalog& ACM \\
\midrule[1pt]
GAE & $80.2\pm1.3$  &$82.9\pm0.3 $&$72.5\pm0.6$\\
AT-L2-GAE &  $\textbf{84.9}\pm0.2$ & $\textbf{84.7}\pm1.4$&$74.2\pm1.7$\\
AT-Linf-GAE& $81.1\pm1.1$& $82.8\pm1.3$&$\textbf{75.3}\pm0.9$\\
\bottomrule[1pt]
\end{tabular}
\vspace{-0.1in}
\end{table}

\section{Understanding Adversarial Training}
In this section, we explore the impact of three hyper-parameters on the performance of GAE and VGAE with adversarial training, i.e.\ the $\epsilon$, $\lambda$ and $T$ in generating $\bm{A}'$ and $\bm{X}'$. These three hyper-parameters are commonly considered to control the strength of regularization for adversarial robustness~\cite{zhang2019theoretically}. Besides, we explore the relationship between the improvements achieved by adversarial training and node degree. 
\subsection{The Impact of $\epsilon$}
The experiments are conducted on link prediction and node clustering tasks based on Cora dataset.  We fix $\epsilon$ to 5e-1 and 1e-3 on adjacency matrix perturbation for $L_{\infty}$ and $L_{2}$ adversarial training respectively when vary $\epsilon$ on attributes perturbation. We fix $\epsilon$ to 1e-3 and 3e-1 on attributes perturbation for $L_{\infty}$ and $L_{2}$ adversarial training respectively when vary $\epsilon$ on adjacency matrix perturbation. 

\begin{figure}
\centering
\vspace{-1in}
\subfloat[Adjacency-Matrix-Perturbation]{\includegraphics[width=0.45\textwidth]{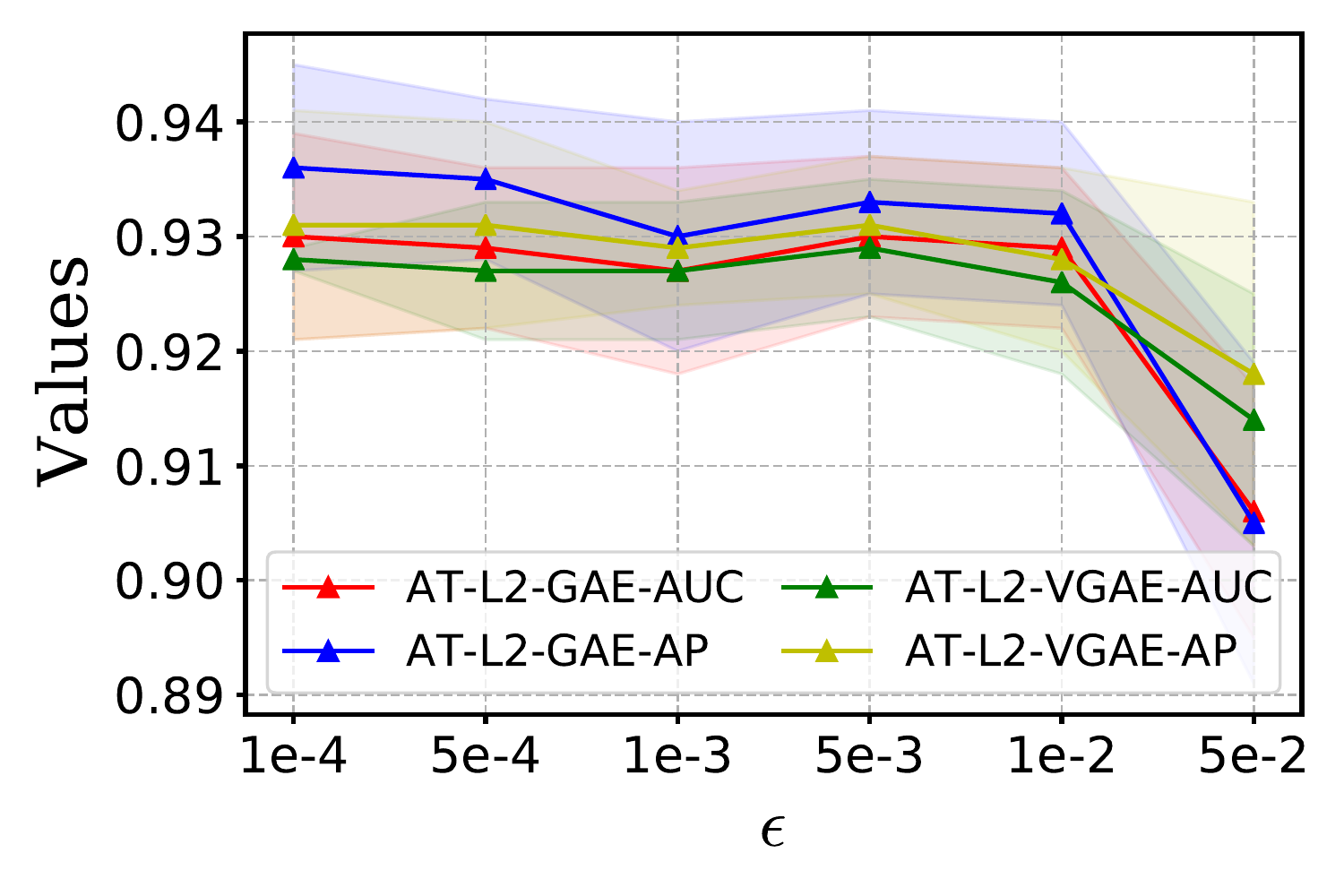}}
\subfloat[Adjacency-Matrix-Perturbation]{\includegraphics[width=0.45\textwidth]{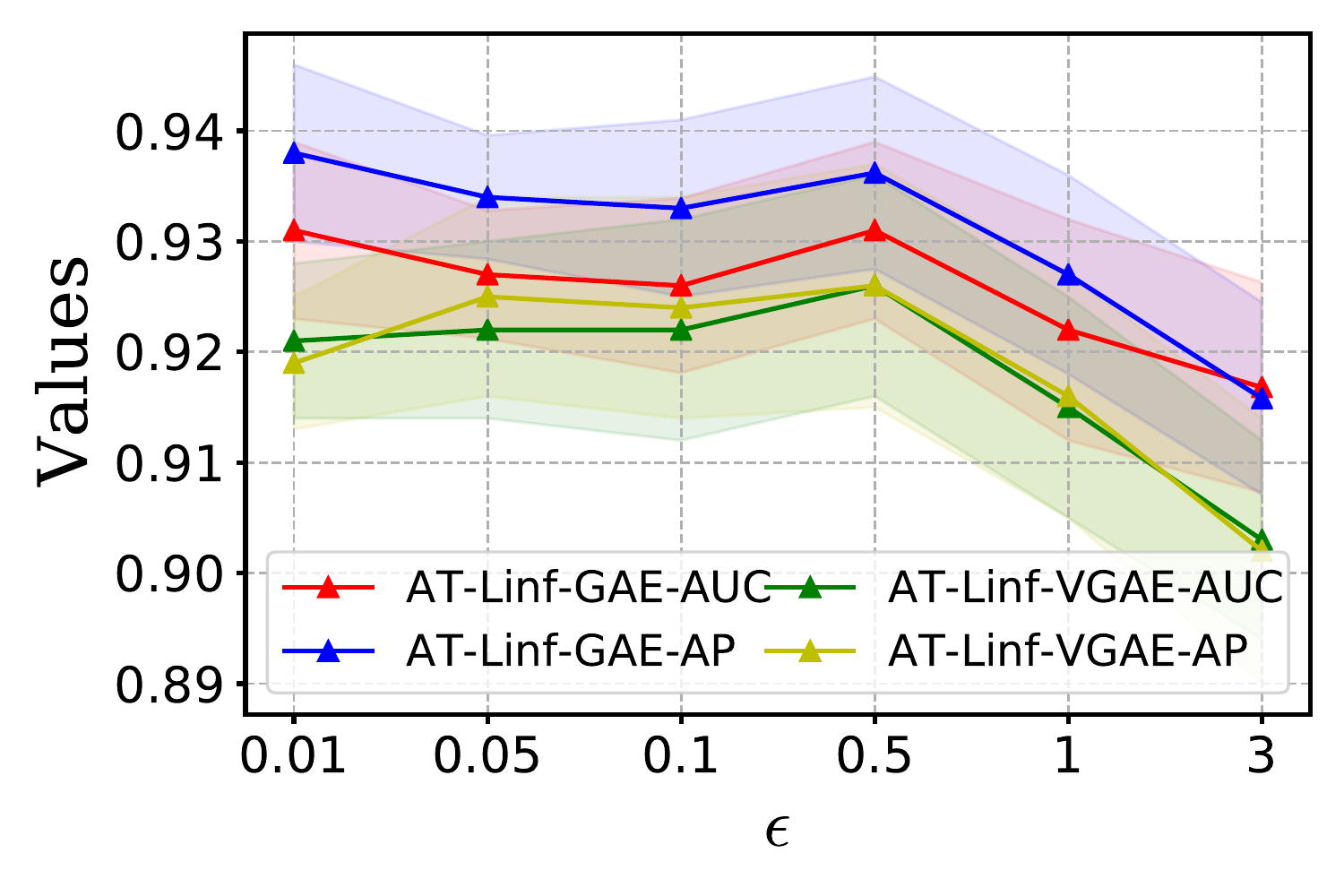}}\\
\subfloat[Attributes-Perturbation]{\includegraphics[width=0.45\textwidth]{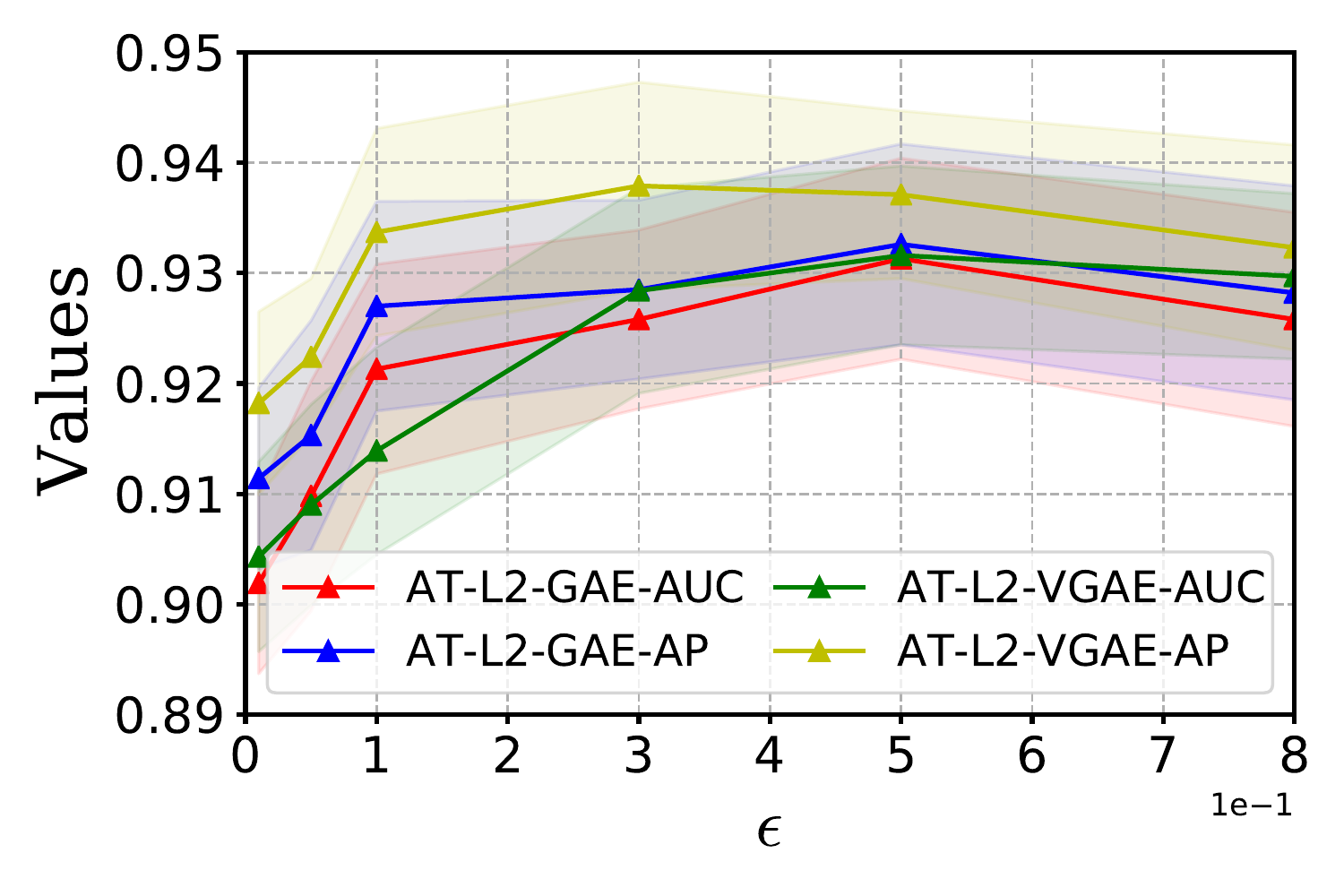}}
\subfloat[Attributes-Perturbation]{\includegraphics[width=0.45\textwidth]{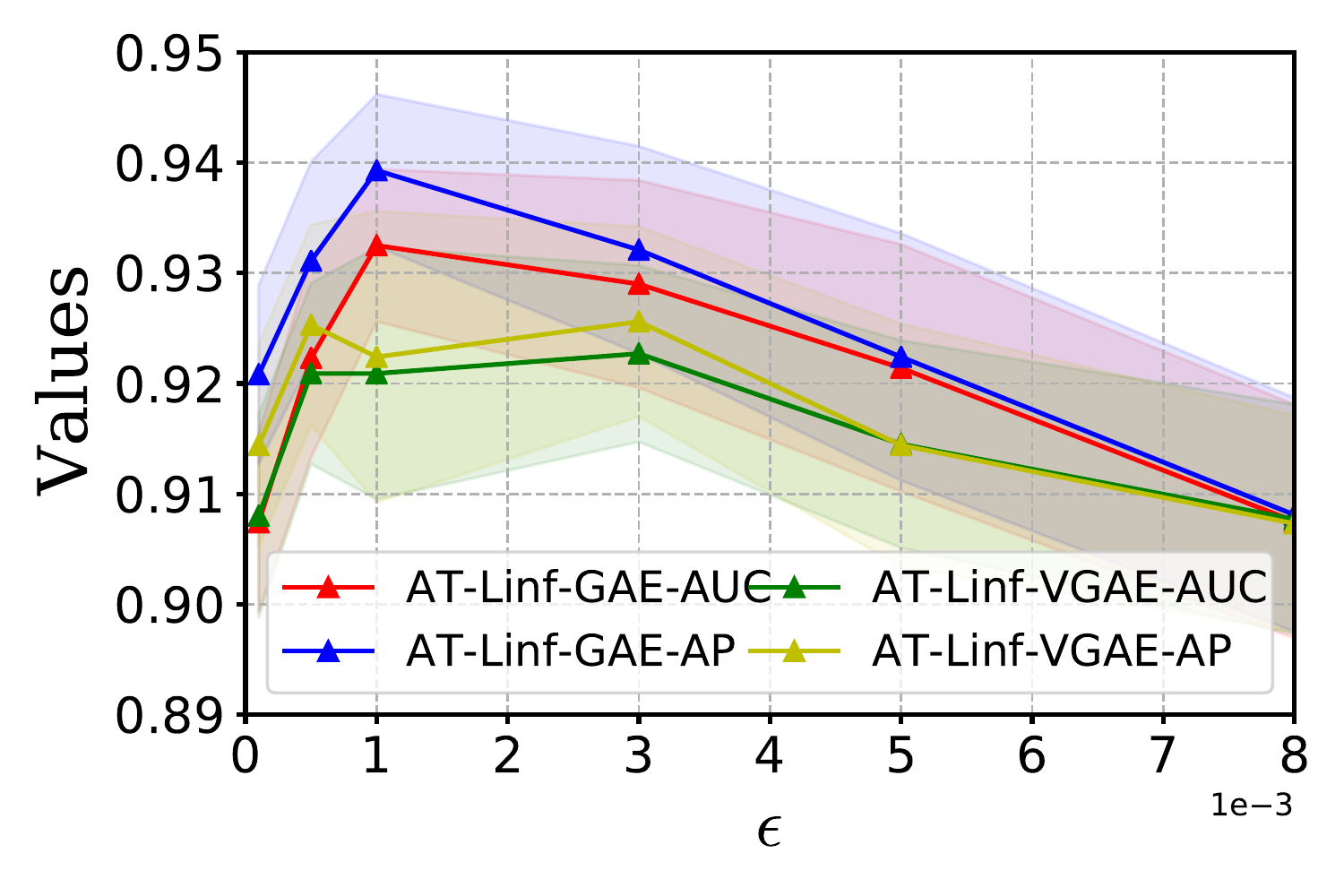}}\\
\subfloat[Adjacency-Matrix-Perturbation]{\includegraphics[width=0.45\textwidth]{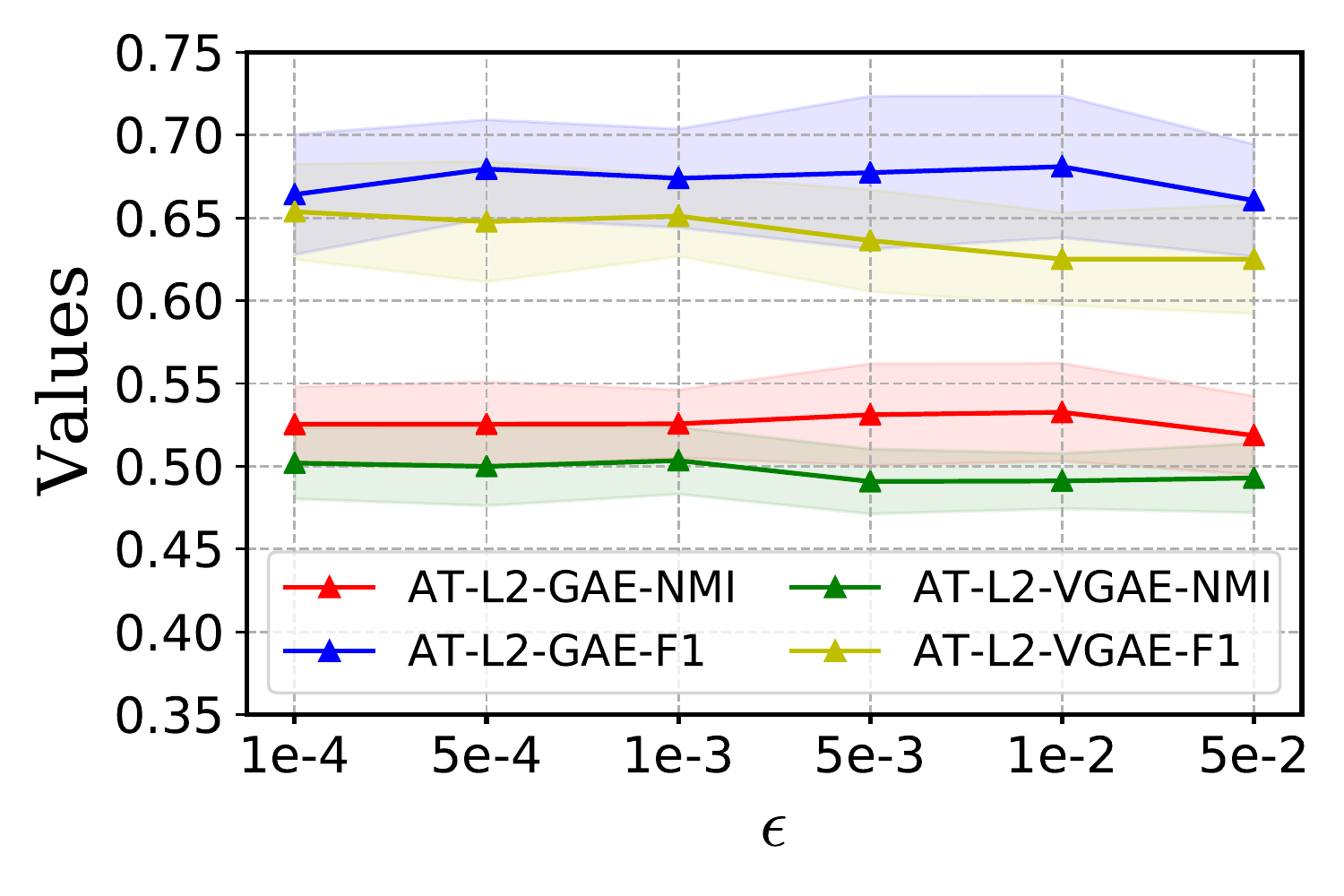}}
\subfloat[Adjacency-Matrix-Perturbation]{\includegraphics[width=0.45\textwidth]{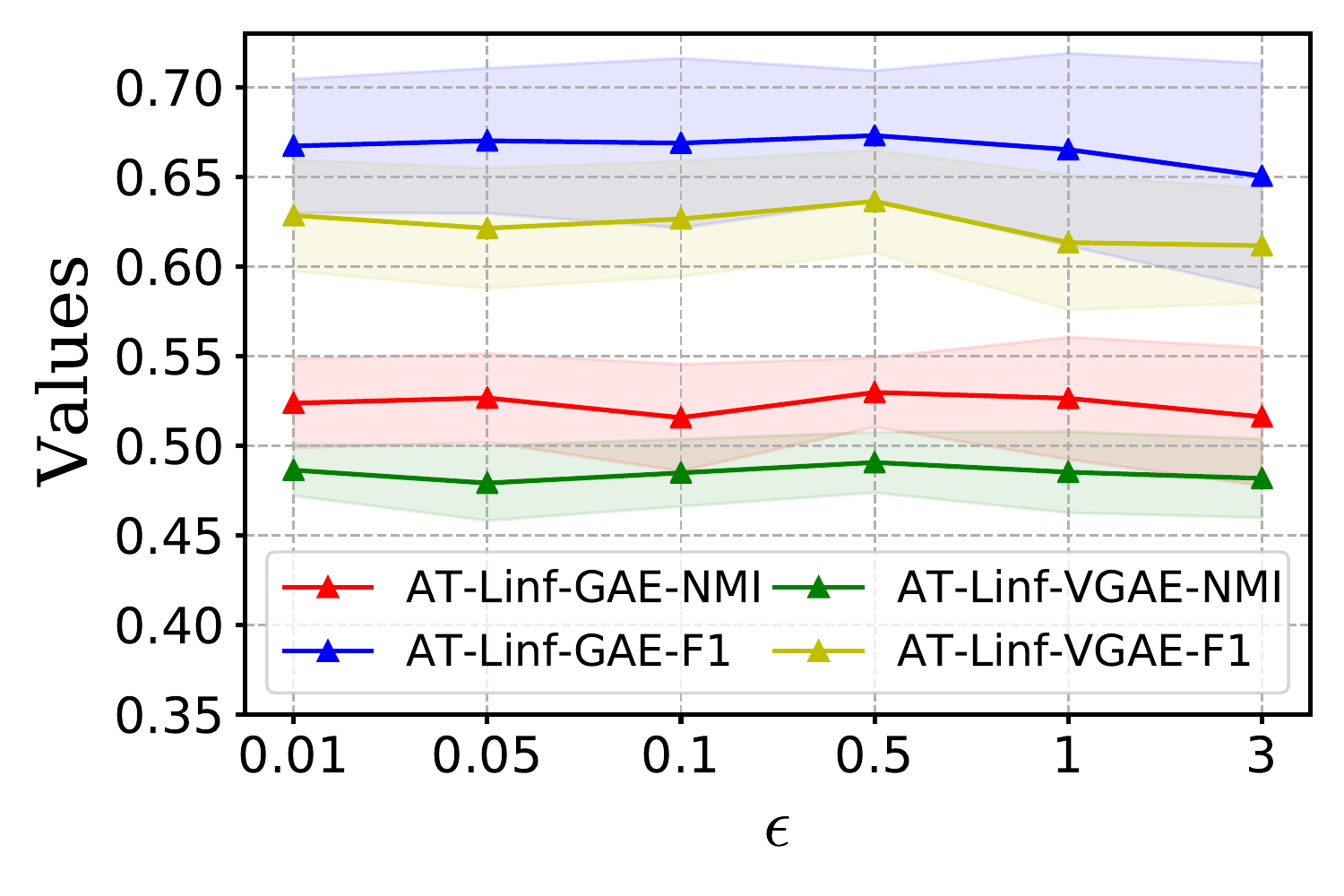}}\\
\subfloat[Attributes-Perturbation]{\includegraphics[width=0.45\textwidth]{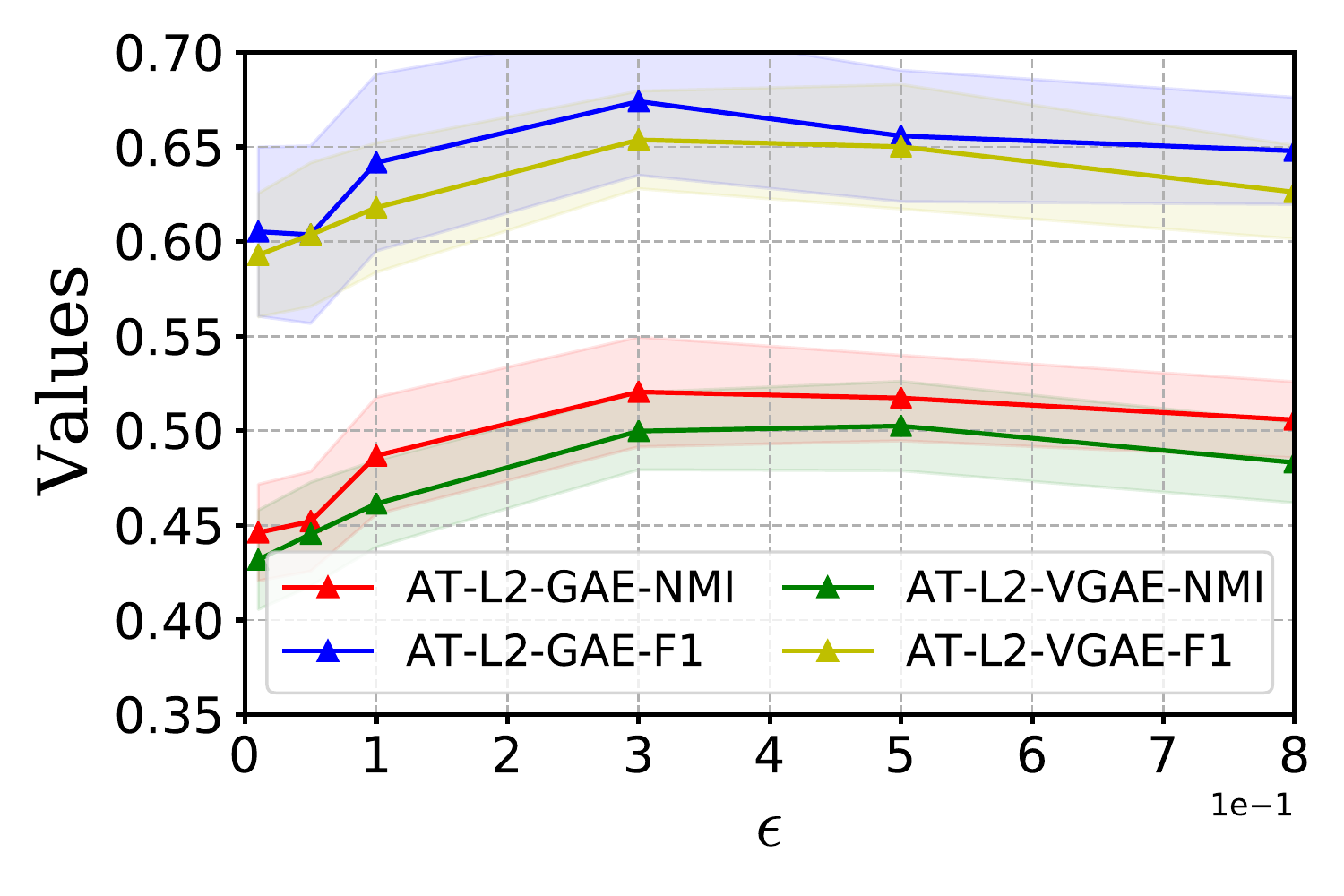}}
\subfloat[Attributes-Perturbation]{\includegraphics[width=0.45\textwidth]{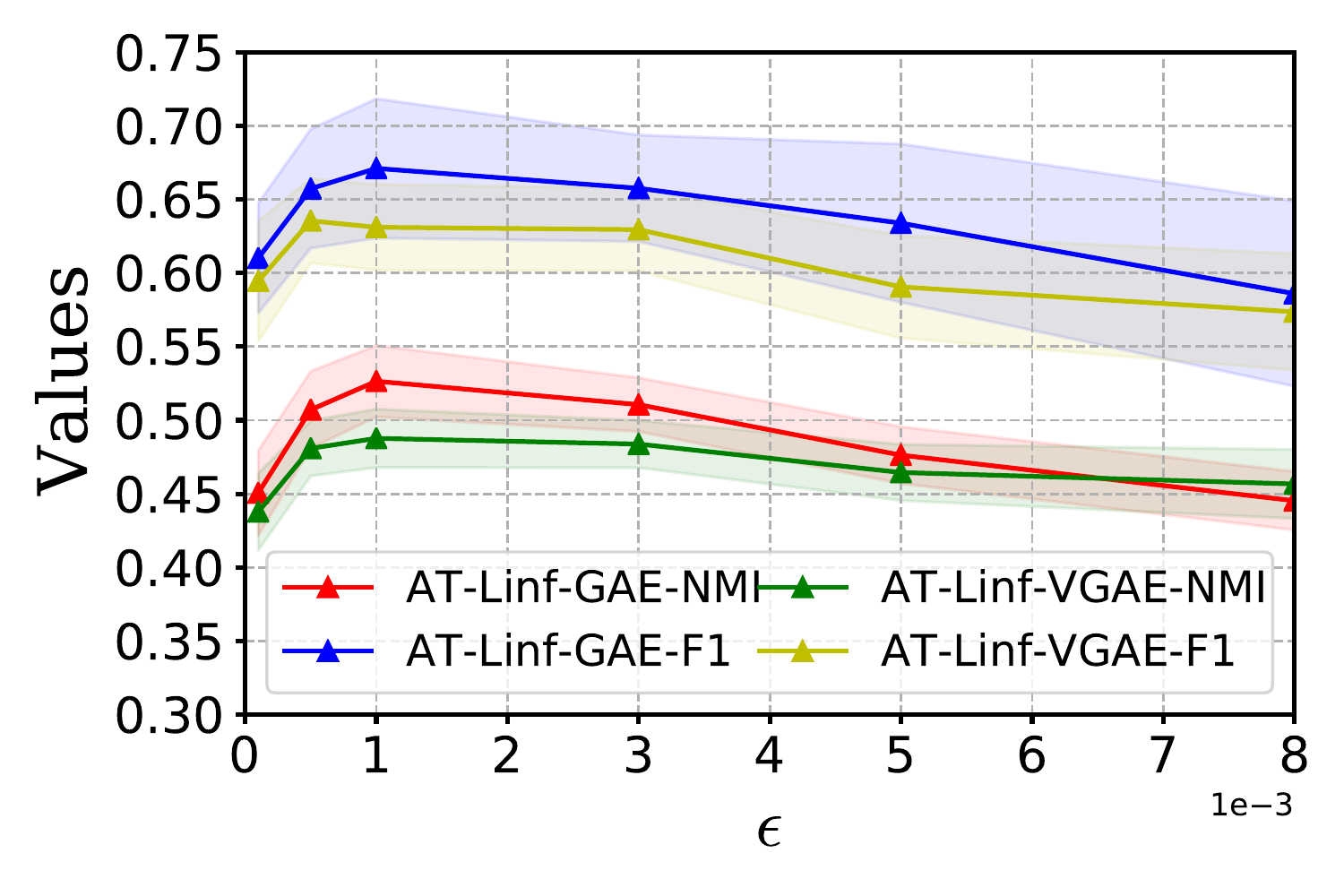}}
\caption{The impact of $\epsilon$ in adjacency matrix perturbation and attributes perturbation. (a)-(d) show AUC/AP values  for link prediction task and (e)-(h) show NMI/F1 values for node clustering task. Dots denote mean values with 30 repeated runs.} \label{analysis_eps}
\vspace{-0.1in}
\end{figure}

The results are showed in Fig.~\ref{analysis_eps}. From Fig.~\ref{analysis_eps}, we can see that the performance are less sensitive to adjacency matrix perturbation and more sensitive to attributes perturbation. Besides, it can be seen that there is an increase and then a decrease trend when increasing $\epsilon$ for attributes perturbation. We conjecture that it is because too large perturbation on attributes may destroy useful information in attributes. Therefore, it is necessary to carefully adapt the perturbation magnitude $\epsilon$ when we apply adversarial training for improving the generalization of model. 

\subsection{The Impact of $T$}
The experiments are conducted on link prediction and node clustering tasks based on Cora dataset. For $L_{2}$ adversarial training, we set $\epsilon$ to 1e-3 and 5e-1 for adjacency matrix perturbation and attributes perturbation respectively. For $L_{\infty}$ adversarial training, we set $\epsilon$ to 1e-1 and 1e-3 for adjacency matrix perturbation and attributes perturbation respectively. We set $\lambda$ to $4$.

\begin{figure}[htb]
\centering
\subfloat[GAE-Link-Prediction]{\includegraphics[width=0.45\textwidth]{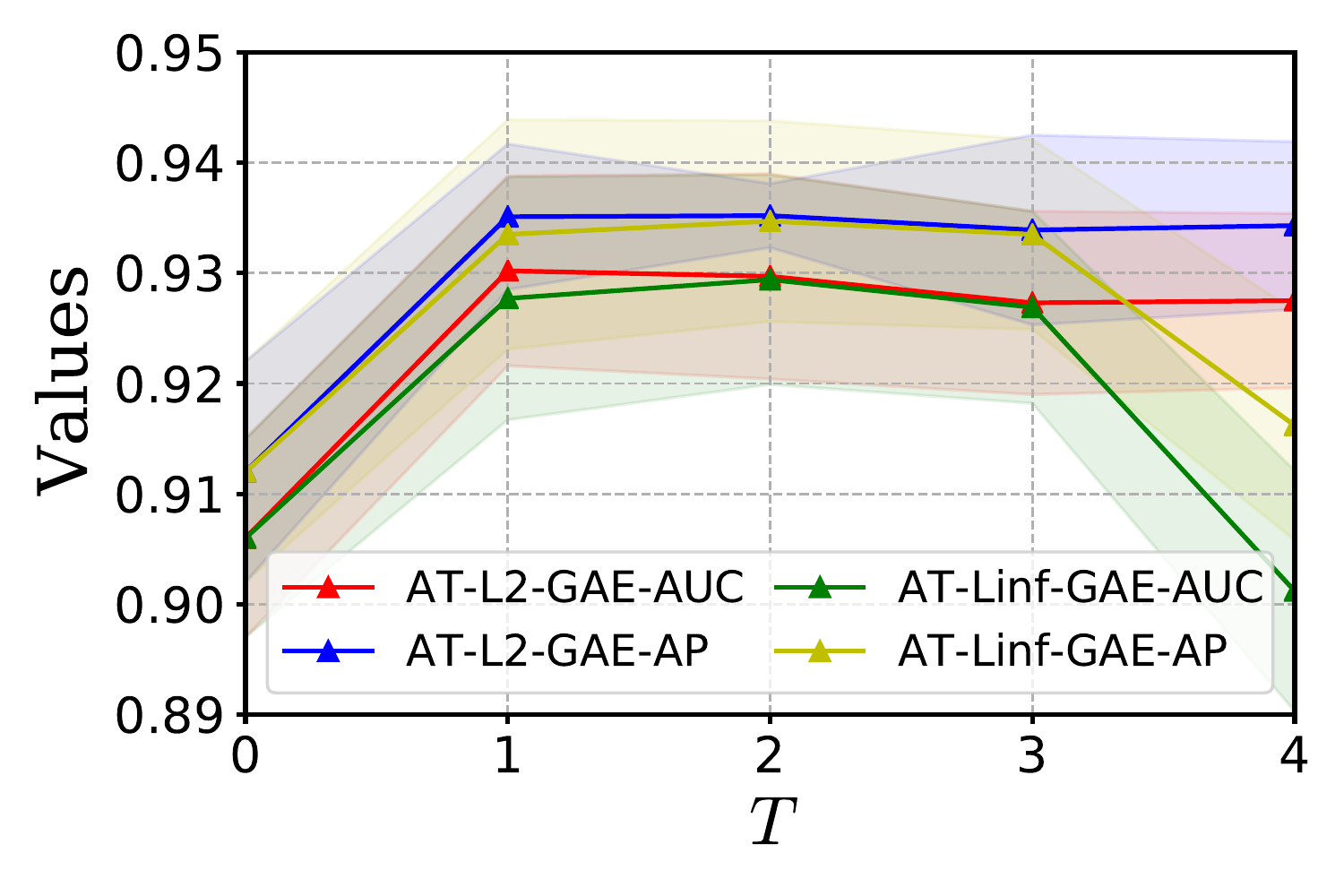}}
\subfloat[VGAE-Link-Prediction]{\includegraphics[width=0.45\textwidth]{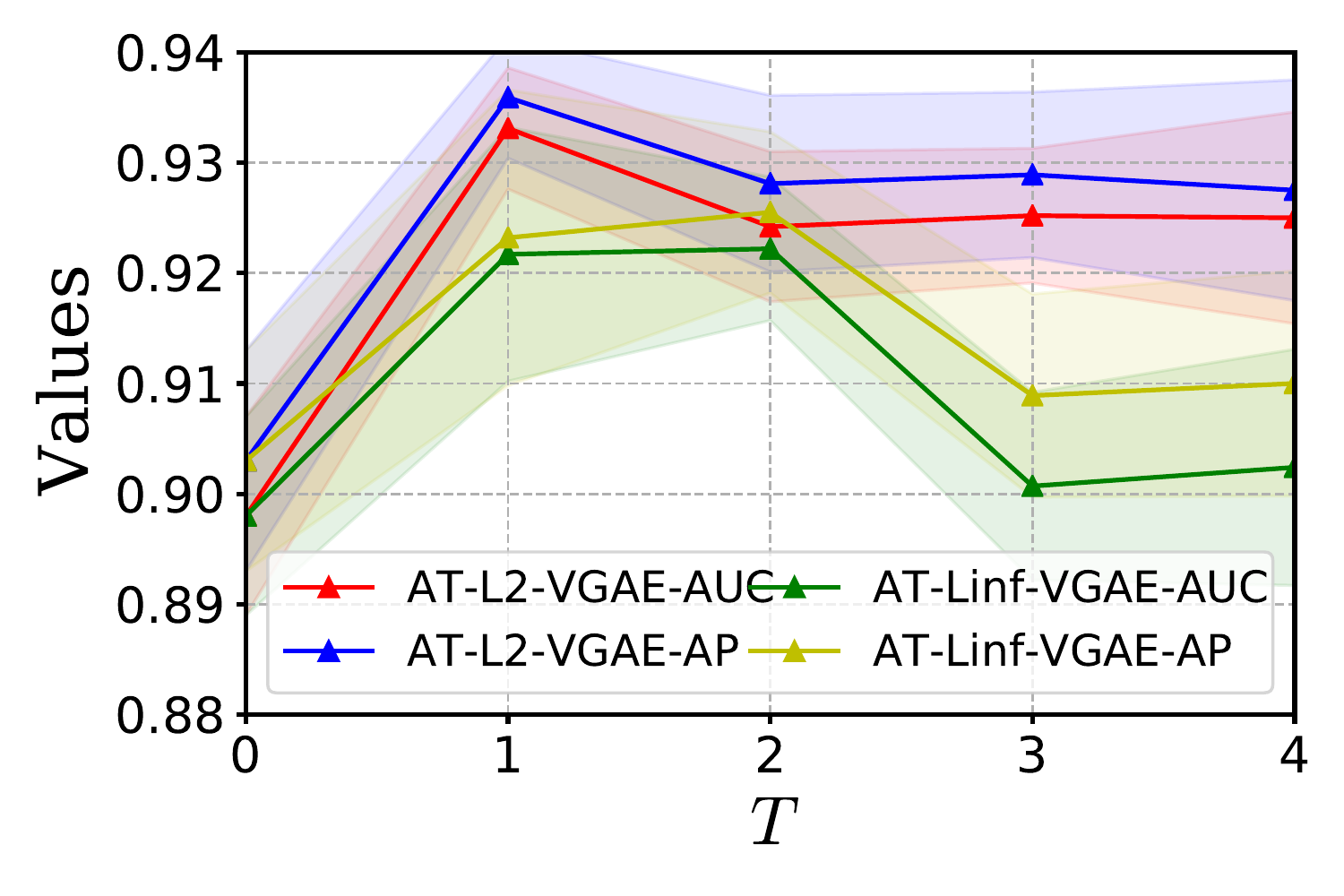}}\\
\subfloat[GAE-Node-Clustering]{\includegraphics[width=0.45\textwidth]{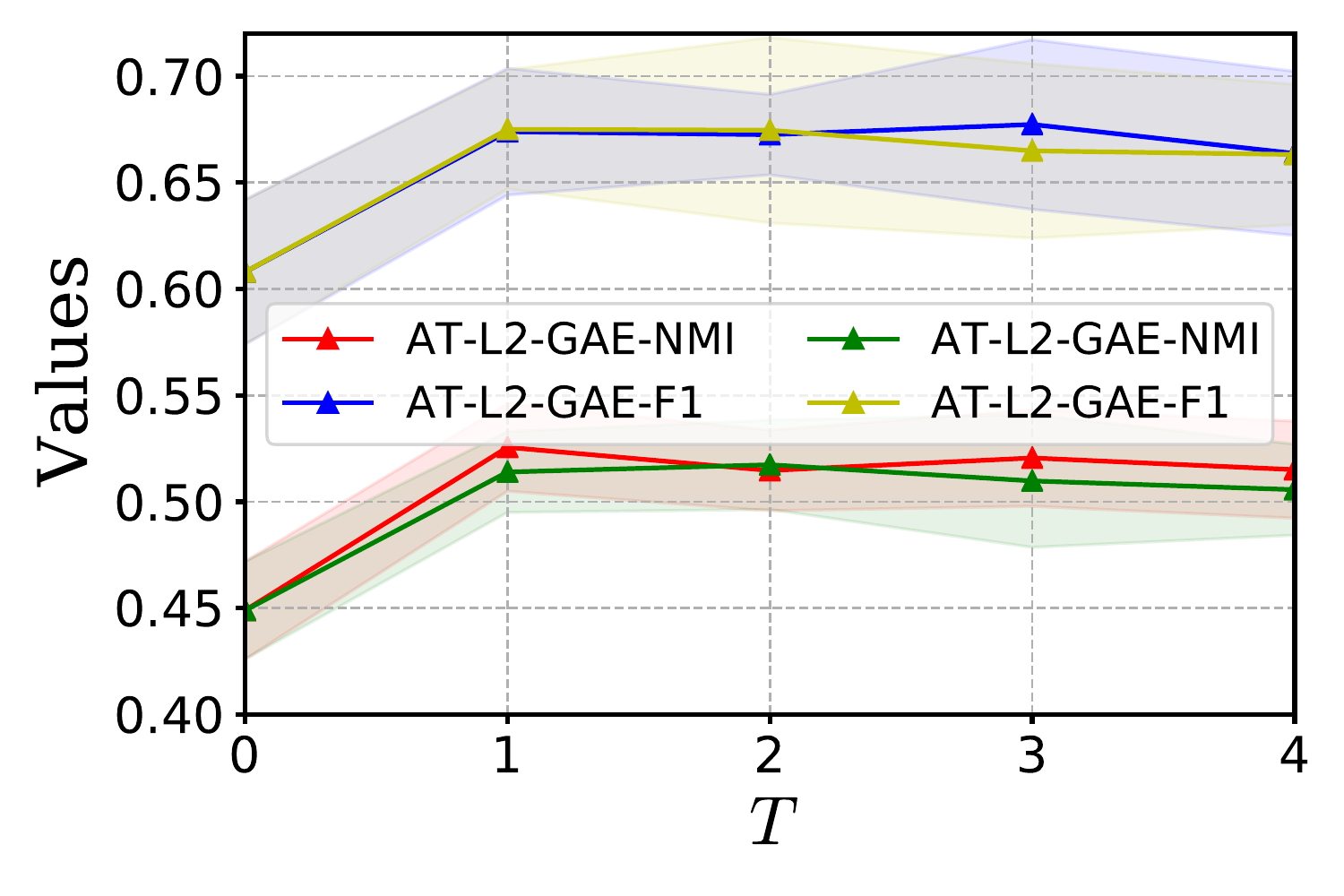}}
\subfloat[VGAE-Node-Clustering]{\includegraphics[width=0.45\textwidth]{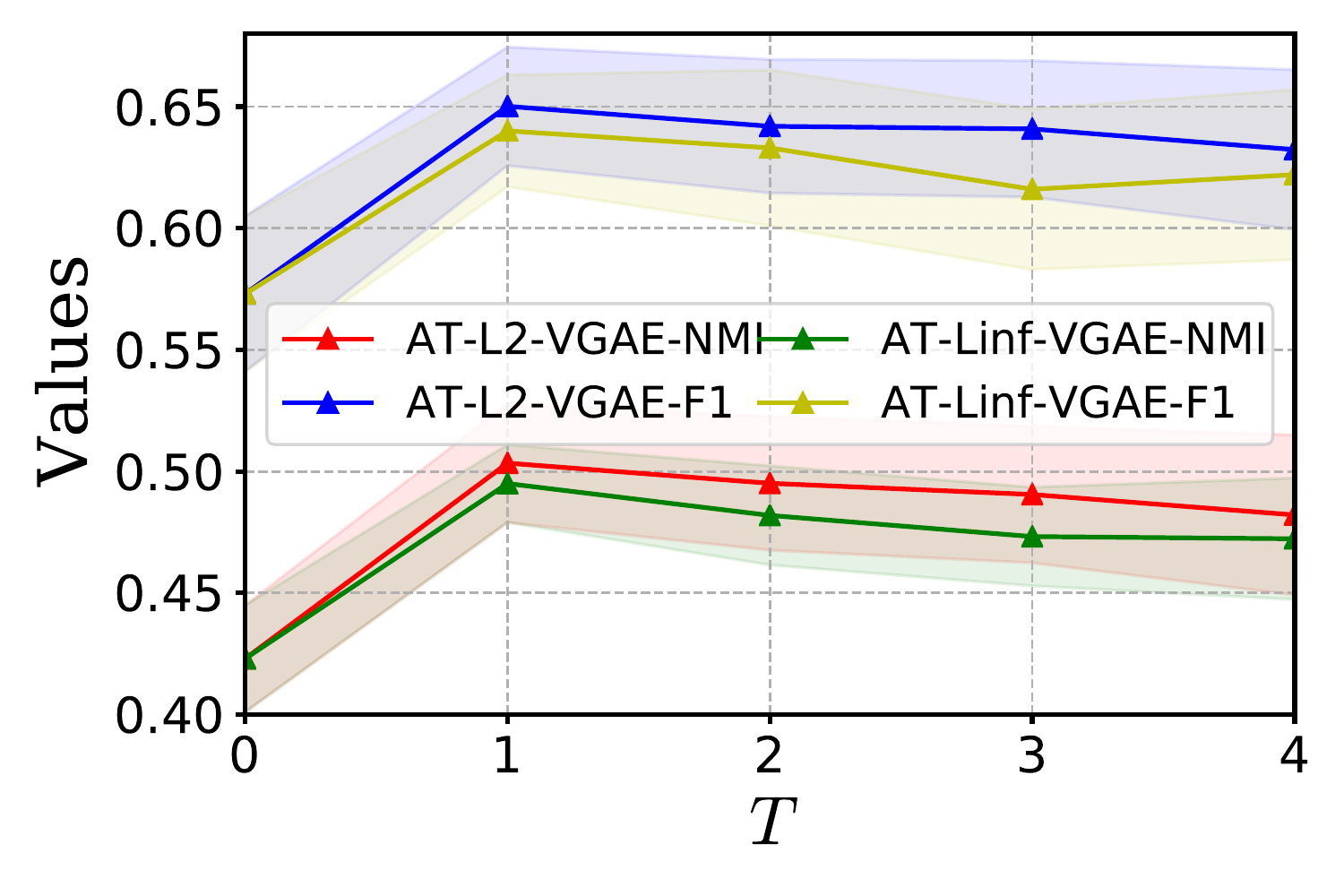}}
\caption{The impact of steps $T$. Dots denote mean AUC/AP values for link prediction task and mean NMI/F1 values for node clustering task. } \label{analysis_steps}
\vspace{-0.1in}
\end{figure}

Results are showed in Fig.~\ref{analysis_steps}. From Fig.~\ref{analysis_steps}, we can see that there is a slightly drop on both link prediction and node clustering tasks  when increasing $T$ from 2 to 4, which implies that a big $T$ is not helpful to improve the generalization of node embeddings learned by GAE and VGAE. We suggest that one step is good choice for generating adjacency matrix perturbation and attributes perturbation in both $L_{2}$ and $L_{\infty}$ adversarial training.

\subsection{The Impact of $\lambda$}
The experiments are conducted on link prediction and node clustering task based on Cora dataset. Likewise, for $L_{2}$ adversarial training, $\epsilon$ is set to 1e-3 and 5e-1 for adjacency matrix perturbation and attributes perturbation respectively. For $L_{\infty}$ adversarial training, $\epsilon$ is set to 1e-1 and 1e-3 for adjacency matrix perturbation and attributes perturbation respectively. $T$ is set to 1. 

\begin{figure}
\centering
\vspace{-0.2in}
\subfloat[Node Clustering]{\includegraphics[width=0.45\textwidth]{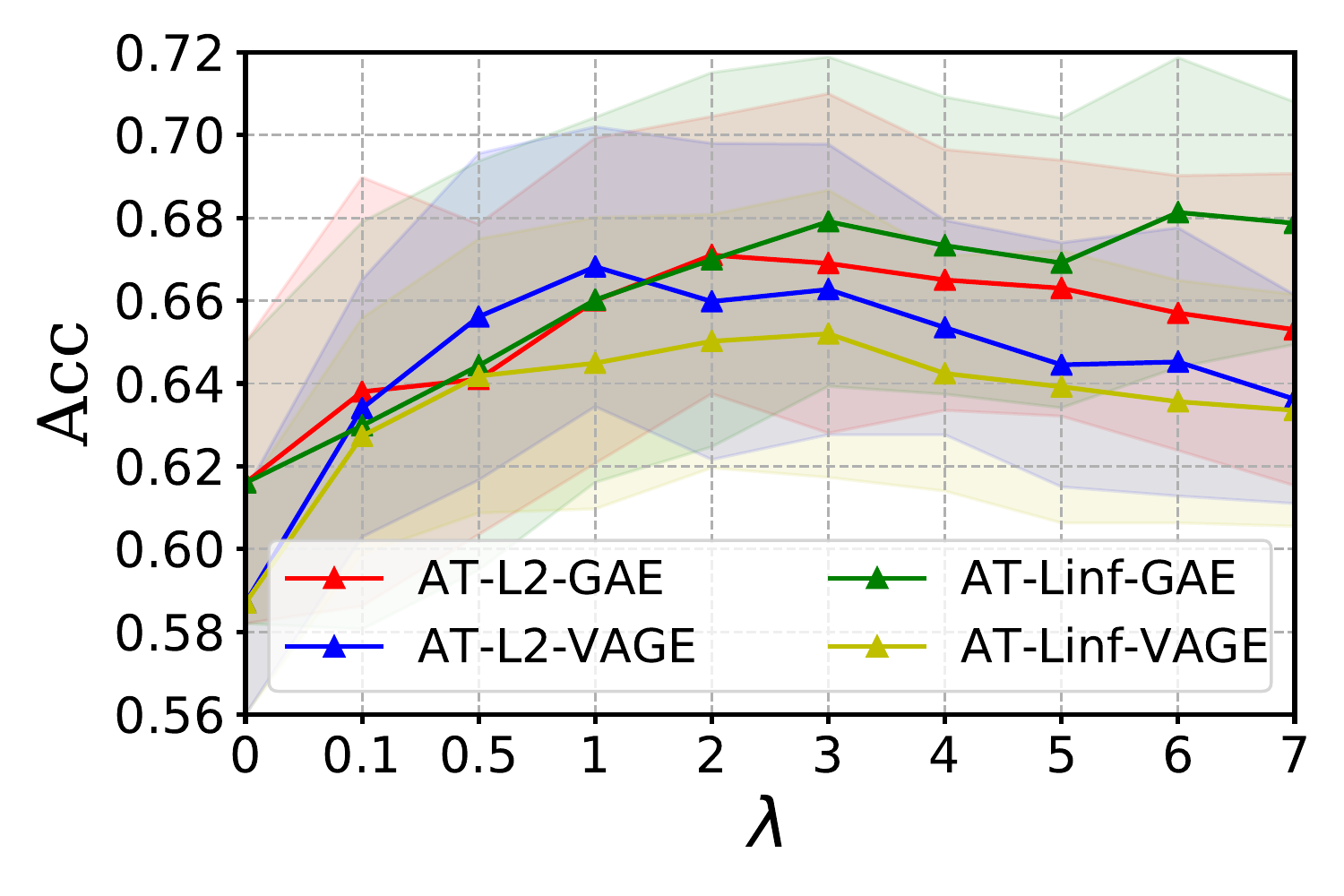}}
\subfloat[Link Prediction]{\includegraphics[width=0.45\textwidth]{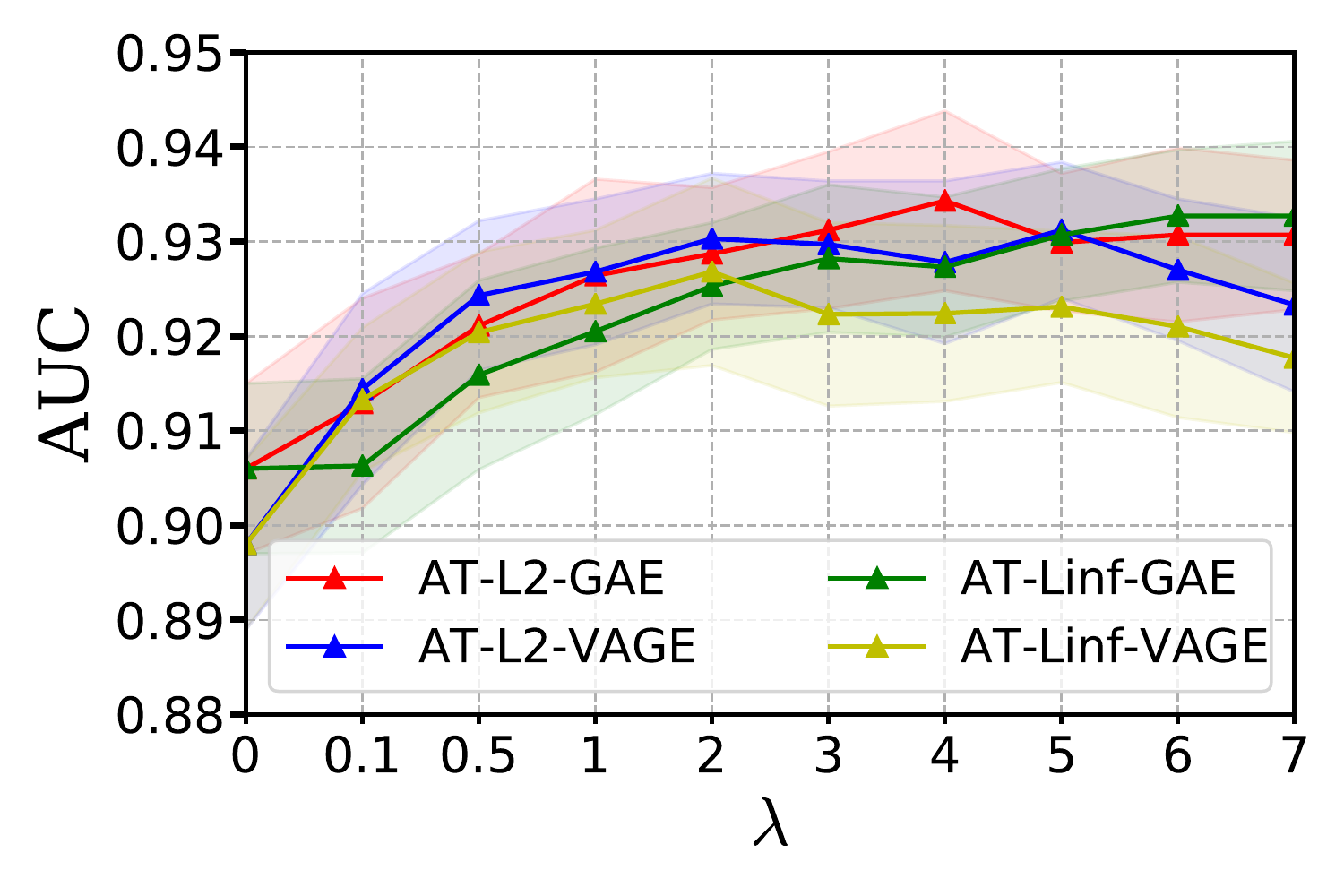}}
\caption{The impact of $\lambda$. $\lambda$ is varied from 0 to 7. Dots denote mean Acc for node clustering task and mean AUC for link prediction task. Experiments are conducted with 30 repeated runs.} \label{analysis_lambda}
\vspace{-0.2in}
\end{figure}

Results are showed in Fig.~\ref{analysis_lambda}. From Fig.~\ref{analysis_lambda}, it can be seen that there is a significant increasing trend with the increase of $\lambda$, which indicates the effectiveness of both $L_{2}$ and $L_{\infty}$ adversarial training in improving the generalization of GAE and VGAE. Besides, we also notice that a too large $\lambda$ is not necessary and may lead to a negative effect in generalization of GAE and VGAE. 

\subsection{Performance w.r.t. Degree}
\begin{figure}
\centering
\vspace{-0.2in}
\subfloat[Cora-GAE]{\includegraphics[width=0.4\textwidth]{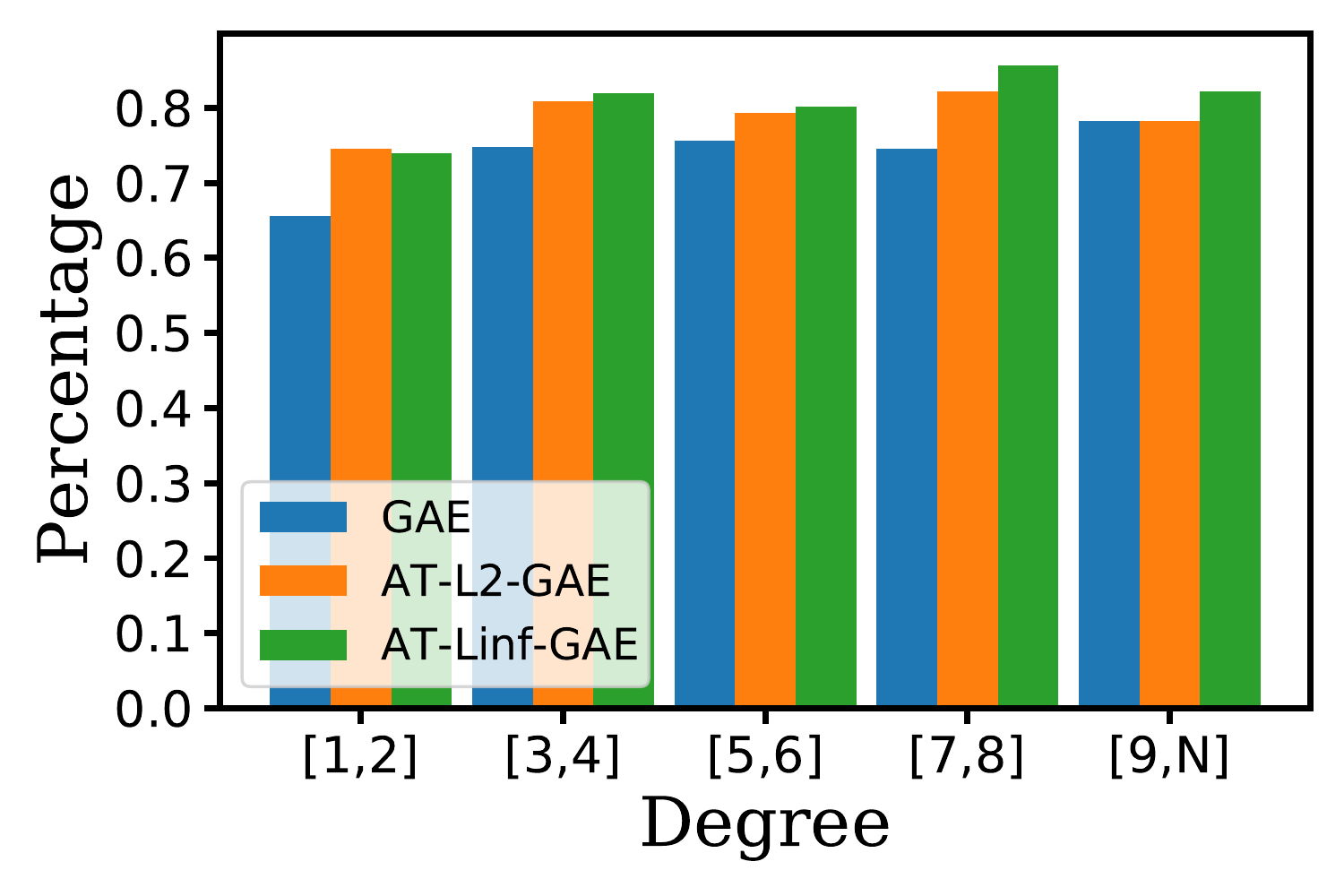}}
\subfloat[Cora-VGAE]{\includegraphics[width=0.4\textwidth]{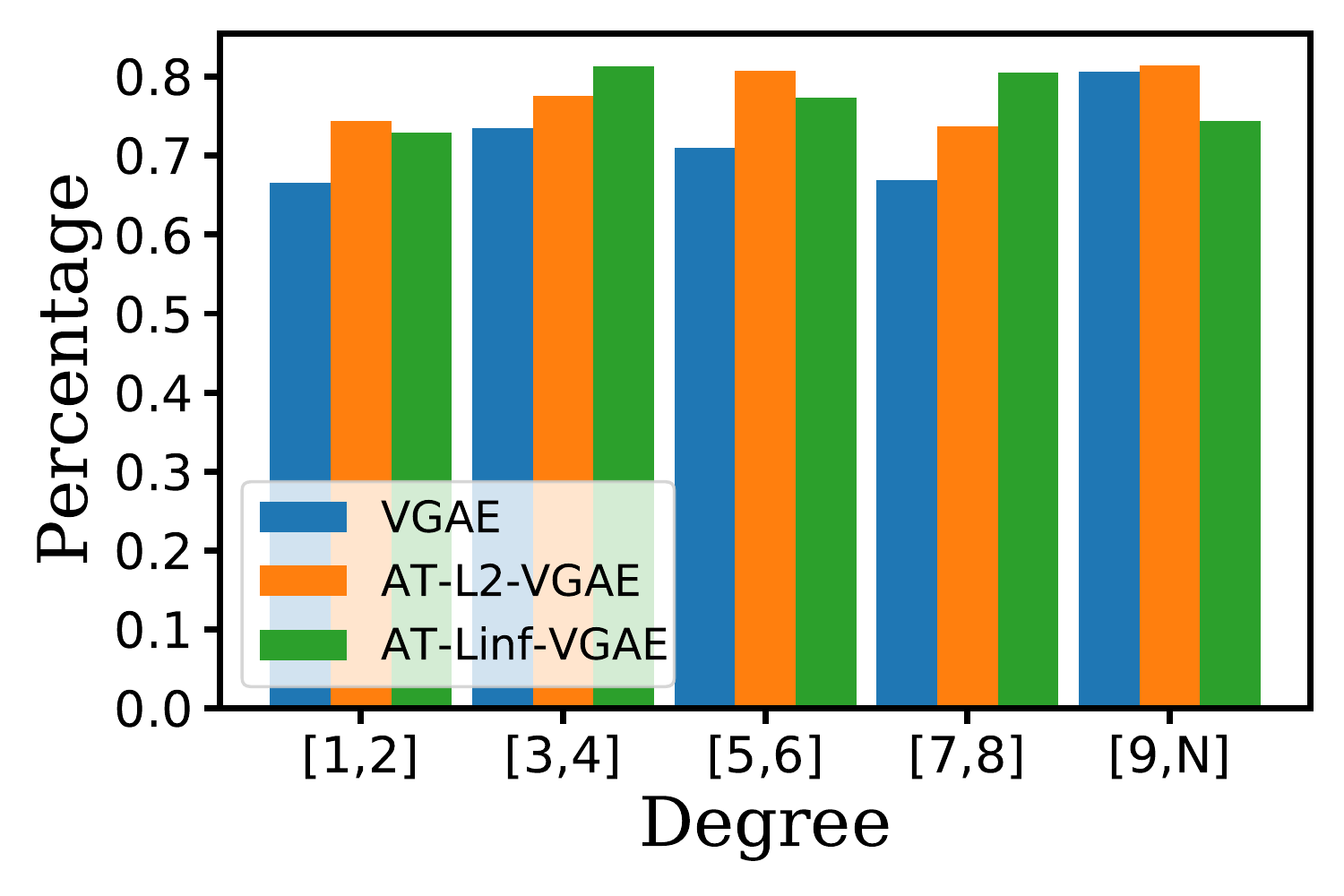}}\\
\subfloat[Citeseer-GAE]{\includegraphics[width=0.4\textwidth]{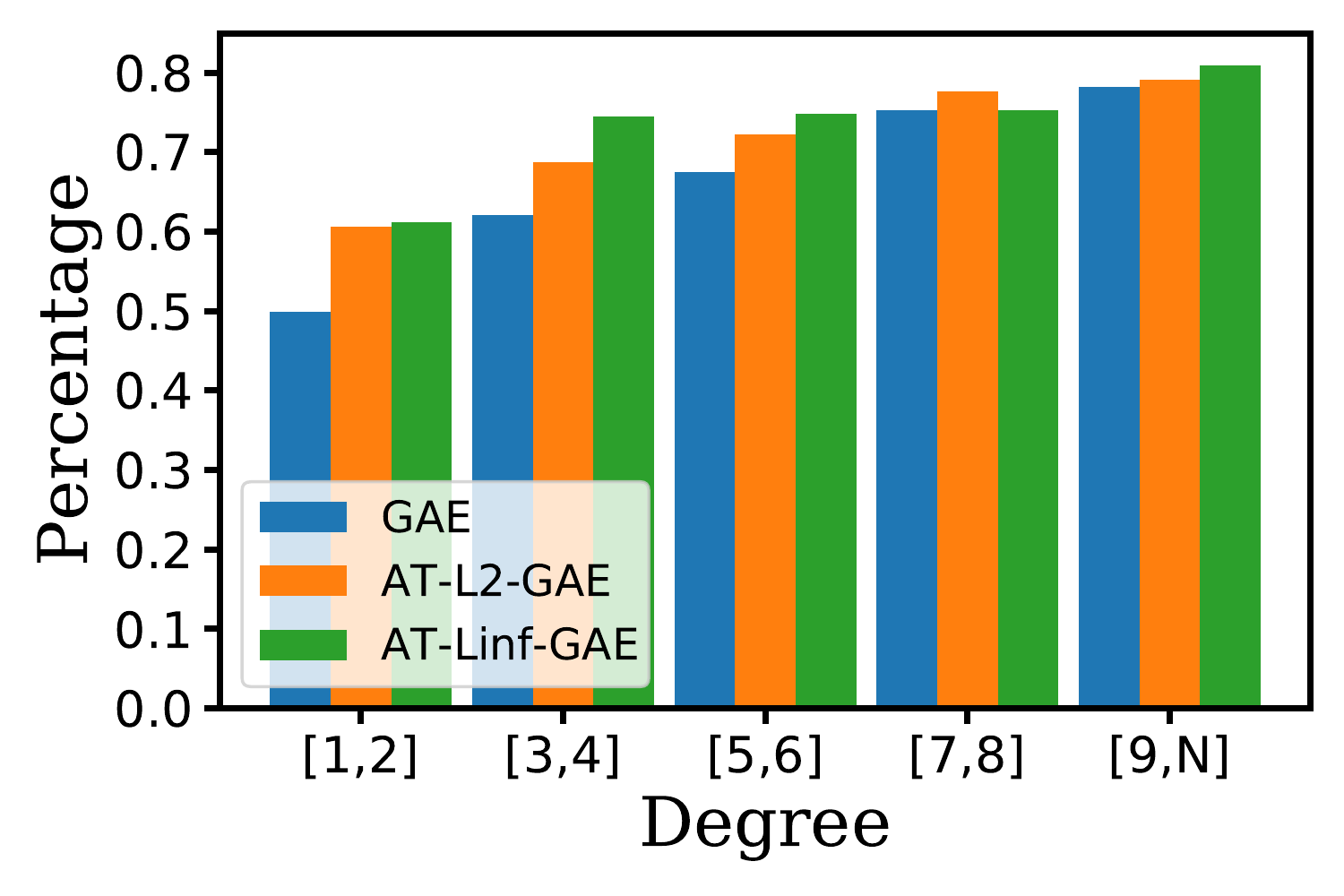}}
\subfloat[Citeseer-VGAE]{\includegraphics[width=0.4\textwidth]{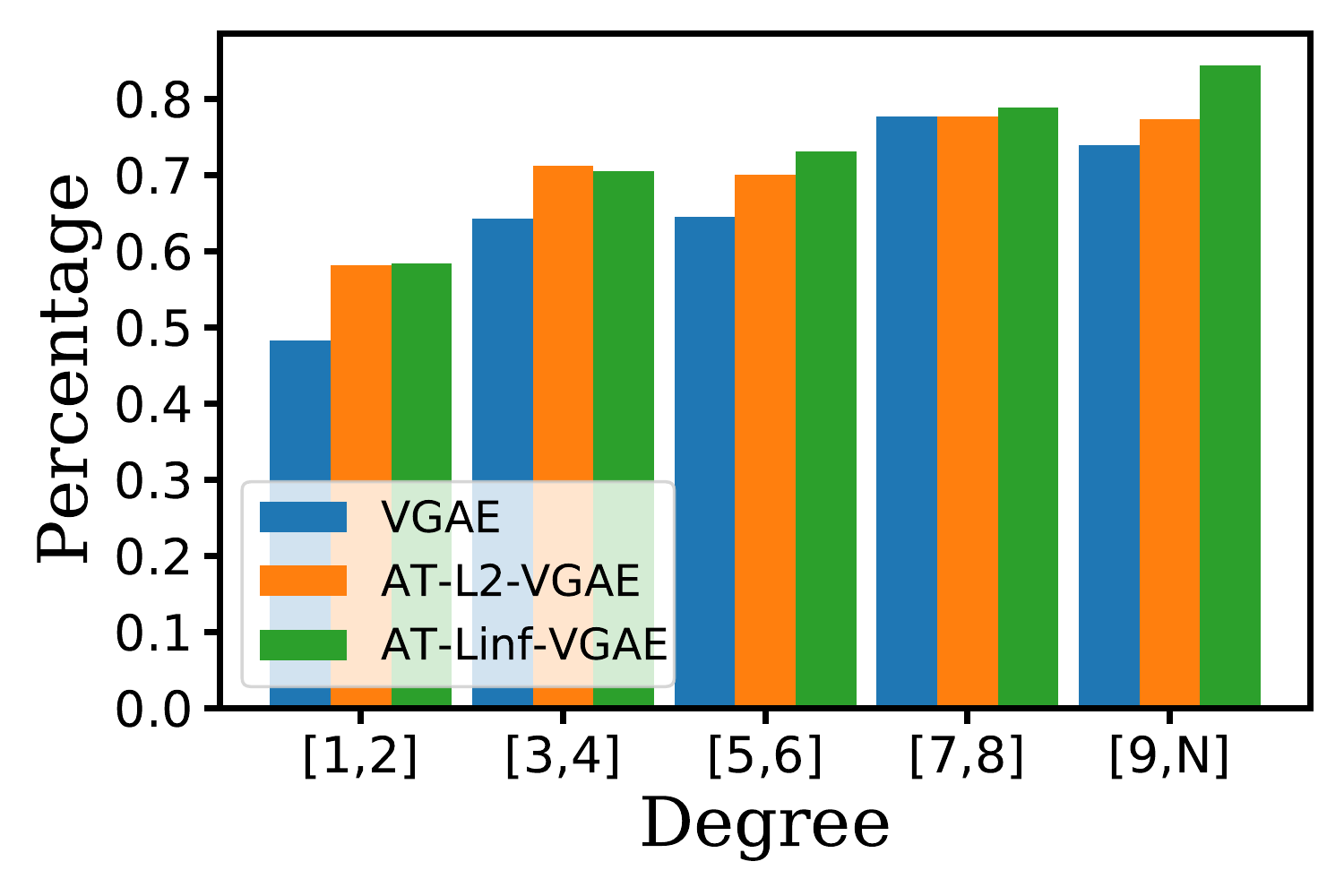}}\\
\caption{Performance of GAE/VGAE and adversarial trained GAE/VGAE w.r.t. degrees in Cora and Citeseer datasets.} \label{Analyis_degree}
\vspace{-0.2in}
\end{figure}

In this section, we explore whether the performance of adversarial trained GAE/VGAE is sensitive to the degree of nodes. To conduct this experiments, we firstly learn node embeddings from Cora and Citeseer datasets by GAE/VGAE with $L_{2}/L_{\infty}$ adversarial training and standard training respectively. The hyper-parameters are set the same as in the Node clustering task.  Then we build a linear classification based on the learned node embeddings. Their accuracy can be found in Appendix. The accuracy with respect to  degree distribution are showed in Fig.~\ref{Analyis_degree}.

From Fig.~\ref{Analyis_degree}, it can be seen seem that for most degree groups, both $L_{2}$ and $L_{\infty}$ adversarial trained models outperform standard trained models, which indicates that both $L_{2}$ and $L_{\infty}$ adversarial training improve the generalization of GAE and VGAE with different degrees. However, we also notice that adversarial training does not achieve a significant improvement on [9,N] group. We conjecture that it is because node embeddings with very large degrees already achieve a high generalization.

\section{Conclusion}
In this paper, we formulated  $L_{2}$ and $L_{\infty}$ adversarial training for GAE and VGAE, and studied their impact on the generalization performance.  We conducted experiments on link prediction, node clustering and graph anomaly detection tasks. The results show that both $L_{2}$ and $L_{\infty}$ adversarial trained GAE and VGAE outperform GAE and VGAE with standard training. This indicates that $L_{2}$ and $L_{\infty}$ adversarial training improve the generalization of GAE and VGAE. Besides, we showed that the generalization performance achieved by the $L_{2}$ and $L_{\infty}$ adversarial training is more sensitive to attributes perturbation than adjacency matrix perturbation, and not sensitive to node degree. In addition, the parameter analysis suggest that a too large $\lambda$, $\epsilon$ and $T$ would lead to a negative effect on the performance w.r.t.\ generalization.

\end{document}